\documentclass[lettersize,journal]{IEEEtran}
\usepackage{amsmath,amsfonts}
\usepackage{algorithmic}
\usepackage{array}
\usepackage{textcomp}
\usepackage{stfloats}
\usepackage{url}
\usepackage{verbatim}
\usepackage{graphicx}
\usepackage{hyperref}
\usepackage{cleveref}
\usepackage{booktabs}
\usepackage{subcaption}
\usepackage{makecell}
\usepackage{multirow}
\usepackage{siunitx}
\usepackage{algorithm}
\usepackage{algorithmic}
\def\BibTeX{{\rm B\kern-.05em{\sc i\kern-.025em b}\kern-.08em
    T\kern-.1667em\lower.7ex\hbox{E}\kern-.125emX}}
\usepackage{balance}

\crefname{figure}{Fig.}{Figs.}
\crefname{equation}{Eq.}{Eqs.}
\crefname{table}{Table}{Tables}
\crefname{section}{Section}{Sections}

\begin{document}

\title{Bridging Quantized Artificial Neural Networks and Neuromorphic Hardware}

\author{Zhenhui Chen, Haoran Xu, Yangfan Hu, Xiaofei Jin, Xinyu Li, Ziyang Kang, Gang Pan, and De Ma%
\thanks{Zhenhui Chen and Haoran Xu are co-first authors.}%
\thanks{Zhenhui Chen and Haoran Xu, Xiaofei Jin, Yangfan Hu, Xinyu Li, Ziyang Kang, Gang Pan are with the College of Computer Science and Technology, Zhejiang University, Hangzhou, China. Email: \{chen\_zhh, xu\_hn, huyangfan, jinxiaofei, xinyu\_li, zykang, ganp\}@zju.edu.cn.}%
\thanks{De Ma is with the College of Computer Science and Technology, Zhejiang University, Hangzhou, China (Corresponding author: made@zju.edu.cn).}
}
\maketitle

\begin{abstract}

Neuromorphic hardware aims to leverage distributed computing and event-driven circuit design to achieve an energy-efficient AI system.
The name ``neuromorphic'' is derived from its spiking and local computing nature, which mimics the fundamental activity of an animal's nervous system.
In neuromorphic hardware, neurons, i.e., computing cores use single-bit, event-driven data (called spikes) for inter-communication, which differs substantially from conventional hardware.
To leverage the advantages of neuromorphic hardware and implement a computing model, the conventional approach is to build spiking neural networks (SNNs).
SNNs replace the nonlinearity part of artificial neural networks (ANNs) in the realm of deep learning with spiking neurons, where the spiking neuron mimics the basic behavior of bio-neurons.
However, there is still a performance gap between SNNs and their ANN counterparts.
In this paper, we explore a new way to map computing models onto neuromorphic hardware.
We propose a Spiking-Driven ANN (SDANN) framework that directly implements quantized ANN on hardware, eliminating the need for tuning the trainable parameters or any performance degradation.
With the power of quantized ANN, our SDANN ensures a lower bound of implementation performance on neuromorphic hardware.
To address the limitation of bit width support on hardware, we propose bias calibration and scaled integration methods.
Experiments on various tasks demonstrate that our SDANN achieves exactly the same accuracy as the quantized ANN.
Beyond toy examples and software implementation, we successfully deployed and validated our spiking models on real neuromorphic hardware, demonstrating the feasibility of the SDANN framework.

\end{abstract}


\section{Introduction}\label{sec:intro}
\IEEEPARstart{W}{ith} the power of modern general-purpose graphics processor, deep learning algorithms have shown impressive capabilities in various cognitive tasks~\cite{dl}. 
However, this performance comes at the cost of immense energy consumption with the traditional computing architecture~\cite{de2023growing}. 
In contrast, the brain of humans and other mammals can process multimodal data and simultaneously understand them at an imperceptible level of energy consumption. 
This naturally inspires research on neuromorphic hardware that mimics the activity of neurons and the nervous system, to achieve low-energy artificial intelligence~\cite{roy2019towards}.

Based on the Network-on-Chip and manycore system~\cite{truenorth,loihi, darwin3}, modern neuromorphic hardware has two main traits:
1) the functional computing unit is arranged as artificial ``neurons'' and ``synapses''. Each neuron transmits information via spikes and integrates the information on the synapse of incoming spikes. 
Here, ``spike'' is the term for the action potential on the axon, which is how a nerve cell communicates with others and receives signals on the synapse in its dendrites. 
2) The neuron can only access the data in the surrounding memory. In other words, data transfer and access are localized~\cite{sparsecoding,lin2018programming}. 
This restricts a single neuron from executing reduce-operations, such as normalizing spike trains in LayerNorm or dot-production attention in the Transformer block~\cite{transformer}. 
The spiking nature makes each neuron use binary values, i.e., $\{0,1\}$ to process information, which simplifies the costly multiply-accumulate operations to merely accumulations by an event-driving circuit.
The localized data access enables us to divide the memory into many distributed computing cores and then bypass the Storage Wall issue. 

Nevertheless, spiking data transfer and localized access also bring unique challenges when we are going to achieve the full potential of neuromorphic hardware. 
Fortunately, lessons from artificial neural networks (ANNs) are helpful in addressing these challenges.
Many previous works have shown that specific types of ANN with neurons formulated as $f(WX+b)$ with input $X$, non-linearity $f$, trainable fixed parameters $W$ and $b$, can be used to build spiking AI models on neuromorphic hardware~\cite{rueckauer2017conversion,sengupta2019going,kim2020spiking}. An intuitive approach is to substitute the non-linearity with a spiking neuron, and then migrate the functionality of deep ANNs to the so-called deep Spiking Neuron Networks (SNNs). 
Hence, many studies have been dedicated to the refinement of the ANN-SNN conversion scheme, whereby the SNN is endowed with the trainable parameters and architecture of its counterpart ANN~\cite{cao2015spiking,diehl2015fast,han2020rmp,deng2021optimal,li2024error}. 
These methods involve post-training calibration of trainable parameters and activation values. However, they also have some disadvantages, such as requiring enormous running rounds, i.e. time steps, to match the performance of ANNs, which significantly impacts running efficiency during inference, incurring additional computation costs and latency. 
Some studies have incorporated special spiking neuron models~\cite{huawei_fp32,hu2023fast}, which are not supported in many neuromorphic processors. 
Another popular approach involves applying Backward-Propagation-Through-Time (BPTT) algorithms to deep SNN, a method referred to as direct training~\cite{wu2018stbp,wu2019direct,zheng2021going,2022mlf}. 
This training approach alleviates the problem of excessive time steps. 
However, the estimated surrogate gradient of the non-differentiable spiking activity of spiking BPTT also has a detrimental effect on the performance of directly trained SNN. 
In addition, the computational overhead and training time of SNNs with BPTT are considerably greater than that of training an ANN \cite{meng2023towards}. This renders directly training large spiking models theoretically possible, but not practically feasible.

\begin{figure*}[t]
    \centering
    \includegraphics[width=\textwidth]{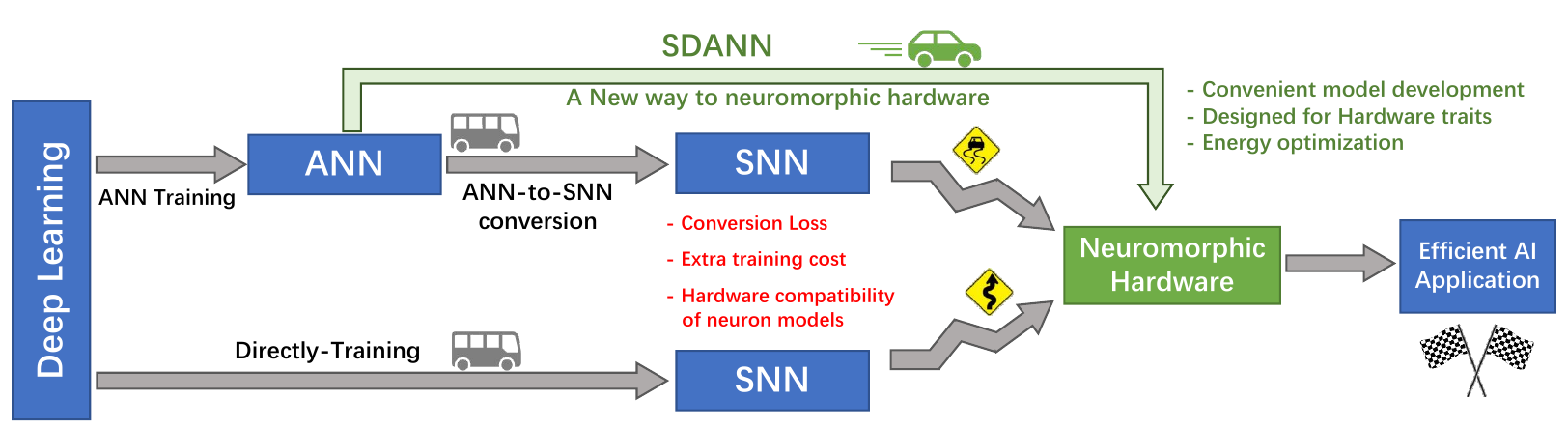}
    \caption{Yet another way from ANN to neuromorphic hardware.}
    \label{fig:big_picture}
\end{figure*}

In this paper, towards the ultimate goal of building energy-efficient neuromorphic systems, we explore a new way to implement computing models on neuromorphic hardware. We propose our Spiking-Driven ANN (SDANN) framework (shown in \cref{fig:big_picture}) that directly maps ANNs to neuromorphic hardware. With this framework, we claim the following contributions.
\begin{itemize}
    \item We propose a Spike-Timing Encoder-Decoder Model (STEM), which can directly deploy quantized ANNs on modern neuromorphic hardware without any weight adaptation and performance damage.
    \item Our framework is hardware-friendly and designed to work in a fixed-point computation environment, facilitating the construction of real neuromorphic applications.  
    \item We propose a sparsification method that reduces the overall number of spikes and energy consumption of the SDANN with minimal performance degradation.
    \item Our experimental results demonstrate the effectiveness of our proposed framework in performance and in reducing energy consumption. Furthermore, we deployed actual models on a \textbf{real} neuromorphic hardware platform for various tasks, advancing in performance and scale (maximum with 12.0M parameters) compared to other SNNs on hardware, as shown in~\cref{tab:SNN on chip}.
\end{itemize}

\begin{table*}[htb]
    \centering
    \caption{The performance of SNNs currently deployed on neuromorphic hardware for various tasks.}
    \label{tab:SNN on chip}
    
    \begin{tabular}{lccccc}
        \toprule
        Work & Datasets & Neuromorphic hardware & \textbf{Parameter} & Precision & Performance \\
        \midrule
        \cite{massa2020efficient}       & CIFAR10                   & Loihi         & 34.1K   & 9 & 77.40\%(Acc.) \\
        \cite{massa2020efficient}       & MNIST                     & Loihi         & 34.1K   & 9 & 98.70\%(Acc.) \\
        \cite{shrestha2021hardware}     & MNIST                     & Loihi         & 41.8K    & 8 & 94.70\%(Acc.)\\
        \cite{shrestha2021hardware}     & Fashion-MNIST             & Loihi         & 41.8K    & 8 & 84.80\%(Acc.)\\
        \cite{shrestha2021hardware}     & CIFAR10                   & Loihi         & 53.8K   & 8 & 62.20\%(Acc.)\\
        \cite{renner2024backpropagation}& MNIST                     & Loihi         & 164K   & 8 & 97.50\%(Acc.)\\
        \cite{frenkel202028}            & MNIST                     & SPOON         & 64.3K    & 8 & 97.50\%(Acc.)\\
        \cite{goltz2021fast}            & MNIST                     & BrainScaleS-2 & 65.4K    & 6 & 96.90\%(Acc.) \\
        \cite{goltz2021fast}            & Yin-Yang dataset          & BrainScaleS-2 & 3.4K    & 6 & 95.00\%(Acc.) \\
        \cite{spilger2023hxtorch}       & Yin-Yang dataset          & BrainScaleS-2 & 960    & 6 & 94.63\%(Acc.) \\
        \cite{huang2023efficient}       & Radar gesture recognition & SpiNNaker 2   & 10.3K   & 8 & 97.60\%(Acc.) \\
        \textbf{Ours}                   & \textbf{ImageNet}         & Darwin3       & \textbf{12.0M}   & 8 & 53.60\%(Acc.) \\
        \textbf{Ours}                   & \textbf{VOC2007}          & Darwin3       & \textbf{5.76M}   & 8 & 50.51(mAP) \\
        \bottomrule
    \end{tabular}
\end{table*}
\section{Preliminary}
\label{sec:preliminary}
\subsection{Quantized ANN}

In the context of this paper, an ANN layer can be formulated as
\begin{align}
    I_i &=\sum_j W_{ij}X_j, \label{eq:MAC} \\
    a_i &= \text{ReLU}(I_i+b_i), \label{eq:relu}
\end{align}
where $W_{ij}$ represents the synaptic weight between the $j$-th neuron in the previous layer and the $i$-th neuron in the current layer, $X_j$ represents the activation value of the $j$-th neuron in the previous layer, and $b_i$ represents the bias term of the $i$-th neuron in the current layer. 
In this context, the non-linearity is implemented using the rectified linear unit (ReLU), which is defined as
$\text{ReLU}(x) = \max\left(x, 0\right)$. A significant number of advanced deep ANN architectures are based on such a combination of linear transformation \cref{eq:MAC} and non-linearity \cref{eq:relu}.

To reduce memory usage and computational cost, neural networks are often quantized by replacing floating-point activations and weights with low-bit integers. Based on the range of activations at each layer, a corresponding set of quantization parameters can be obtained to map the floating-point values to $n$-bit integers. 
For each variable in \cref{eq:MAC,eq:relu}, we have two major constants: the scaling factor $S$ and the zero-point $Z$ defined by  
\begin{align}
    S&=\frac{r_{\max}-r_{\min}}{q_{\max}-q_{\min}}, \label{scale}\\
    Z&=\text{round}\left(q_{\max}-\frac{r_{\max}}S\right).\label{zero point}
\end{align}
Here, $r_{\max}$ and $r_{\min}$ are the observed maximum and minimum values of the activation outputs from \cref{eq:MAC}. $q_{\max}$ and $q_{\min}$ are the two ends of the target integer width ($q_{\max}=127$ and $q_{\min}=-128$ if we assign 8-bit integer for example).
Then we use the two constants to get
\begin{align}
    q &= \text{round}(\frac rS+Z) \label{quantize}, \\
    r &\approx S\cdot(q-Z) \label{dequantize},
\end{align}
where $r$ represents a floating-point value, $q$ is the integer obtained after quantization, $S$ is the scaling factor, and $Z$ is the zero-point.
So \cref{eq:MAC} can be rewritten as:
\begin{equation}
\begin{split}
    S_a\cdot(\tilde{a}_i-Z_{a}) \approx {} &
    \text{ReLU}\Bigg(\sum_j S_w\cdot(\tilde{W}_{ij}-Z_w)S_x\cdot \\
    & (\tilde{X}_j-Z_x) + S_b \cdot (\tilde{b}_i-Z_b)\Bigg),
\end{split}
\label{eq:quantizeMAC_ann}
\end{equation}
where $\tilde{a}_i$, $\tilde{W}_{ij}$, $\tilde{x}_j$, $\tilde{b}_i$ correspond to the quantized variable of $a_i$, $W_{ij}$, $x_j$, $b_i$ in \cref{eq:MAC}. $S_a$ and $Z_a$ represent the scale factor and zero-point for variable $a$, respectively. In this paper, we adopt the uniform quantization~\cite{jacob2018quantization} with a symmetric quantization range, which means that the zero-point $Z$ is fixed to zero. This choice is motivated by its simplicity of implementation and hardware compatibility, especially for neuromorphic systems with fixed-point arithmetic. Compared with asymmetric or non-uniform quantization schemes, symmetric quantization introduces fewer hardware overheads and allows direct integer operations, which aligns with the design constraints of our target platform.
With this quantization setup, we have:
\begin{align}
    S_a\tilde{a}_i &\approx \text{ReLU}\left(\sum_jS_w\tilde{W}_{ij}S_x\tilde{X}_j+S_b\tilde{b}_i\right) \notag \\
    \tilde{a}_i& \approx \max\left\{ \frac{S_wS_x}{S_a}\sum_j\tilde{W}_{ij}\tilde{X}_j+ \frac{S_b}{S_a}\tilde{b}_i,0\right\}. \label{eq:quantizeMAC}
\end{align}
Since ${S_wS_x}/{S_a}$ and ${S_b}/{S_a}$ are constants, the computation of the weighted sum and bias can be fully implemented using integer arithmetic. The corresponding multiplication can be achieved via fix-point multiplication and bit shift.

\subsection{Neuromorphic Hardware and Models for Neuromorphic Hardware}
\begin{figure*}[t]
    \centering
    \includegraphics[width=0.8\textwidth]{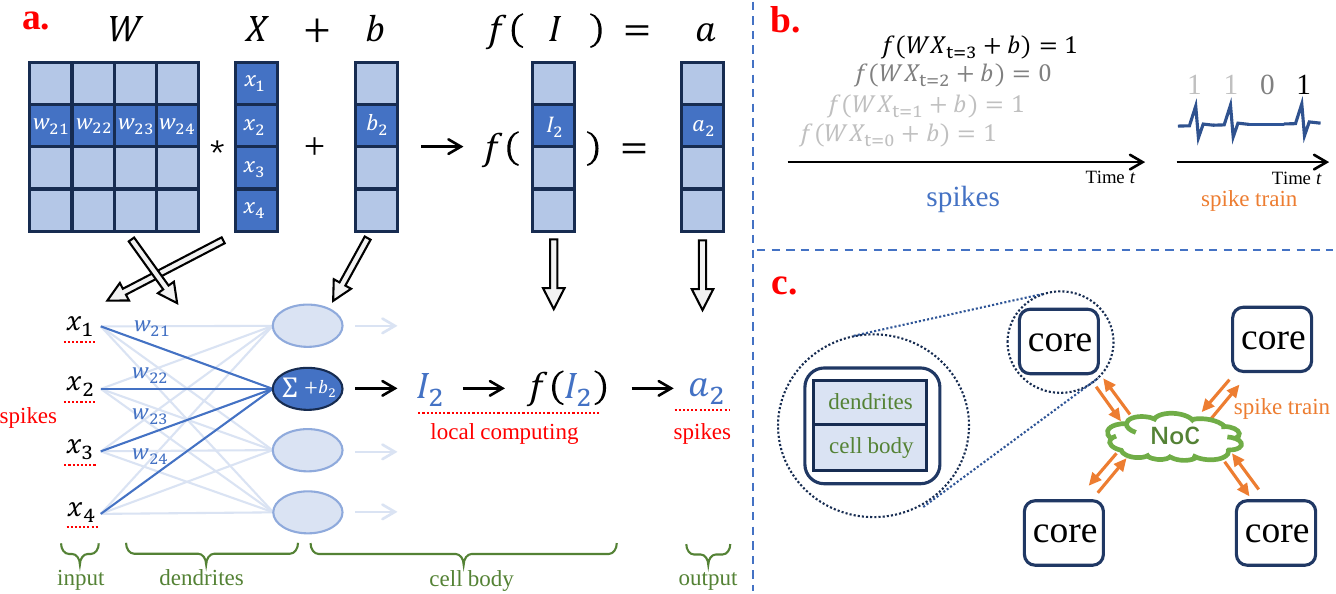}
    \caption{a) From weighted sum to dendrites and neurons. b) From spikes to spike train. c) A simple neuromorphic hardware architecture with NoC.}
    \label{fig:demo}
\end{figure*}
Neuromorphic hardware organizes compute cores and memory differently compared to conventional Von Neumann architectures.
Computation in neuromorphic hardware is carried out by neurons and dendrites, resembling the nervous system in animals.
Parameters and neuron states are stored within these structures.
These signals emulate action potentials in biological neurons, known as spikes, which excite subsequent neurons.
Unlike general-purpose computers, neuromorphic hardware lacks a globally accessible memory.
Neurons are distributed and cannot directly access each other's states or parameters, unless they are interconnected via 1-bit signal links.
Hence, building a hardware compatible model means not only adding spiking datapaths, but also ensuring that the model components can be transformed into the ``dendrites'' and ``cell body'', as shown in \cref{fig:demo}a.
Obviously, Multiply–Accumulate (MAC) operations as \cref{eq:MAC} can be implemented into many connections between consecutive layers.
Note that the output activation value, denoted by \cref{eq:relu}, is not inherently a single-bit signal, which contradicts the hardware design. The output of these operations must be in spikes. As mentioned in \cref{sec:intro}, ANN2SNN and direct training methods replace the ReLU component of \cref{eq:relu} with simplified spiking neuron models, such as the leaky integrate-and-fire (LIF) model, to restrict input/output to 1-bit signals.
In addition to that, neural states can only be updated by its corresponding distributed compute core, as shown in \cref{fig:demo}c. So element-wise operations are only permitted at the same level of ReLU in \cref{eq:relu} as the non-linearity $f$, as another requirement of the model. 
Note that the input/output process can be iterated in multiple turns, in a process called the temporal neuron dynamic, which is termed {\it running in multiple time steps} in the SNN domain.
This behavior mimics the temporal dynamics of biological neurons, i.e. accumulate spikes in the {\it historical} membrane potential and fire accordingly. 
The final output is a spiking train that contains spikes from each time step, as shown in \cref{fig:demo}b. Our SDANN method leverages the temporal information in the spike train to directly deploy a quantized ANN on hardware.

\section{The SDANN framework}\label{sec:method}
\begin{figure*}[t]
    \centering
    \includegraphics[width=\textwidth]{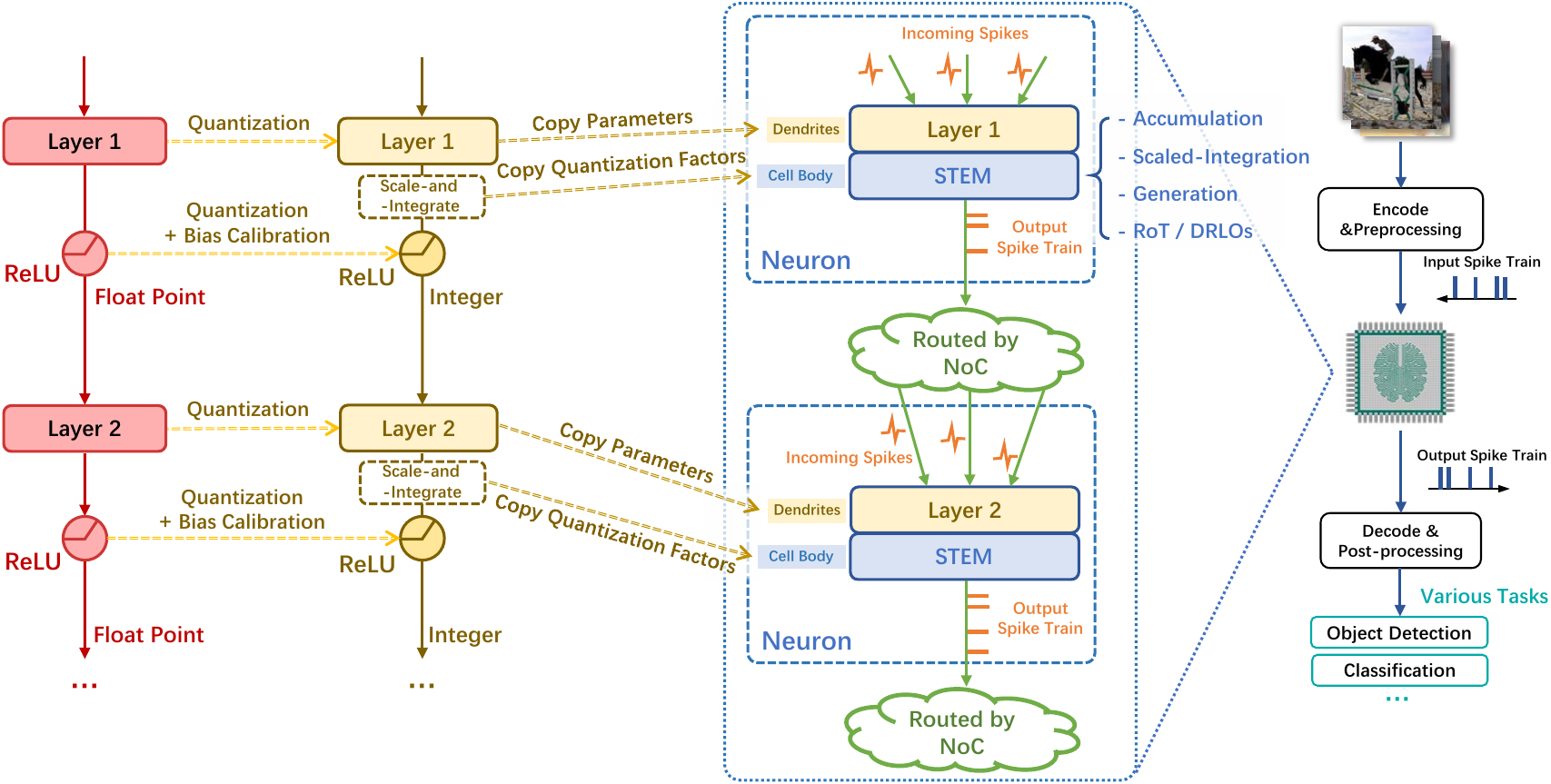}
    \caption{The workflow of SDANN framework, from ANN quantization to neuromorphic hardware implementation.}
    \label{fig:framework}
\end{figure*}

The workflow of the SDANN framework is illustrated in \cref{fig:framework}, consisting of two major steps:
\textbf{1)} Adapting the uniform quantization method to quantize the weights and activations. 
During quantization, factors related to \textbf{scaled integration} and \textbf{bias calibration} are statistically gathered. 
The scaled integration method is employed to mitigate the risk of overflow during weight accumulation, while bias calibration adjusts the bias to meet hardware data width requirements. 
\textbf{2)} Modeling the quantized ANN with STEM involves four distinct steps:
\begin{itemize}
\item \textbf{Accumulation}: The STEM will receive the spike train and interpret the input as a signed integer as a {\it spike decoder}.
\item \textbf{Scaled integration}:  We rescale the intermediate result during the accumulation process, which may exceed the capacity of the hardware,and result in overflow. 
\item \textbf{Generation}: The STEM will generate spike trains according to the results from {\it Accumulation}, where the STEM will {\it encode} the results into a spike train.
\item (Optional) \textbf{RoT \& DRLOs}: Two optional spike sparsity methods are offered: the Round-off Truncation (RoT) and the Discarding Redundant Low-order Spikes (DRLOs). These methods are employed to reduce the number of transmitted spikes as well as the energy cost. 
\end{itemize}
Subsequently, the quantized ANN is prepared for implementation on neuromorphic hardware and for various tasks.
It is noted that no extra training or calibration procedures are implemented in terms of trained parameters. Furthermore, an inference pipeline has been developed to reduce the time step required for inferences. The following subsections provide detailed introductions to these components.

\subsection{Bias calibration}
\label{sec:bias calibration}
\begin{figure*}[t]
    \centering
    \begin{subfigure}[b]{0.45\textwidth}
        \centering
        \includegraphics[width=\textwidth]{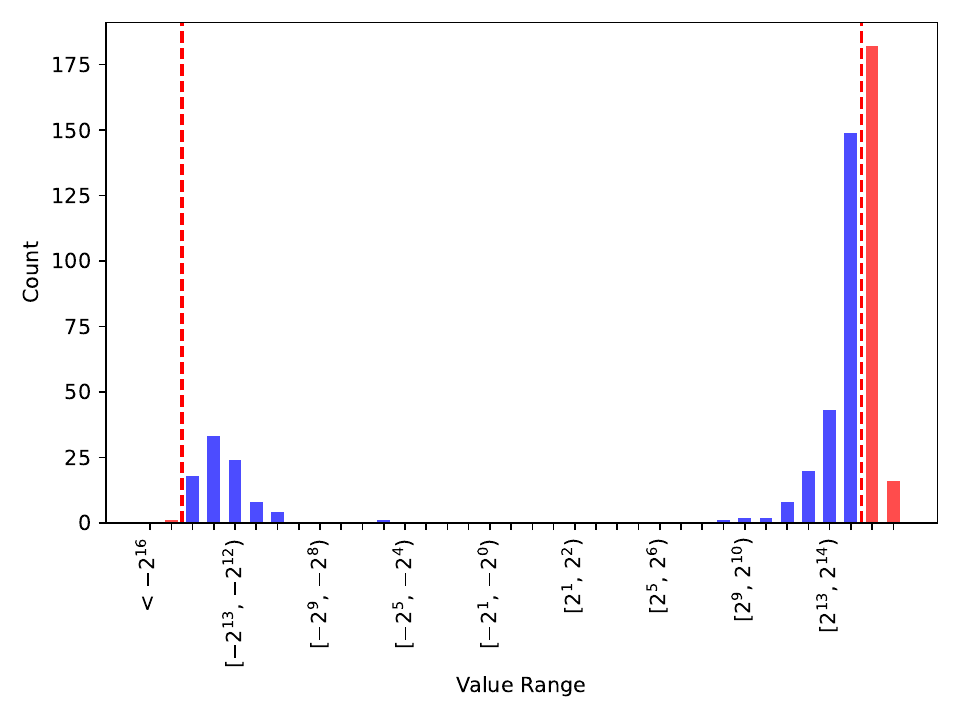}
        \caption{Without bias calibration}
        \label{fig:conv2_wo_bias_cali}
    \end{subfigure}
    \hfill
    \begin{subfigure}[b]{0.45\textwidth}
        \centering
        \includegraphics[width=\textwidth]{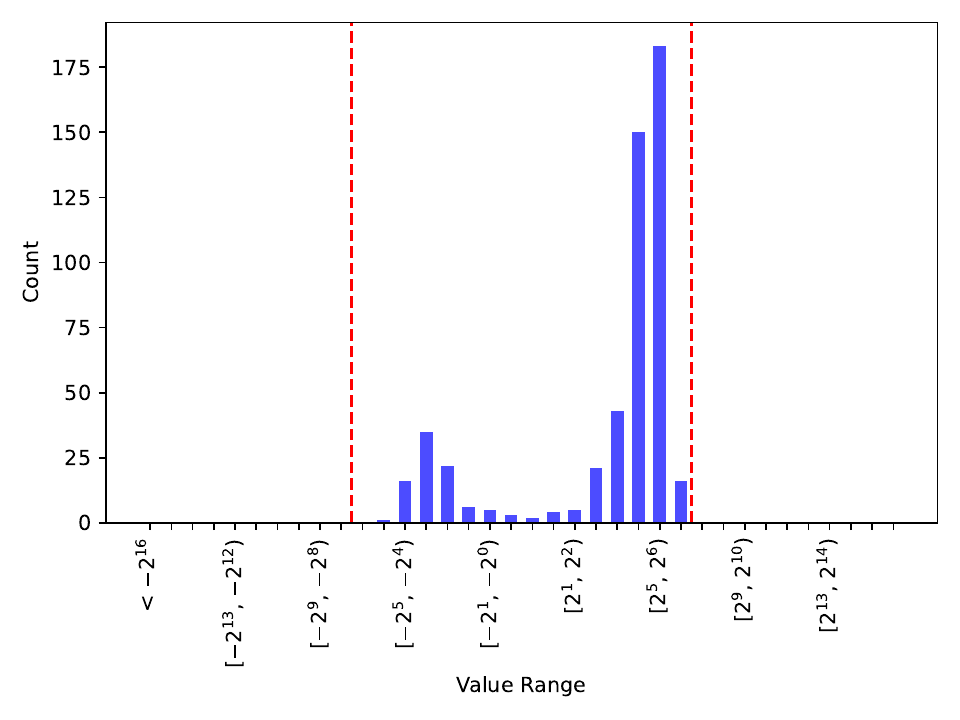}
        \caption{With bias calibration}
        \label{fig:conv2_w_bias_cali}
    \end{subfigure}
    \caption{The distribution of quantized bias values across logarithmically scaled intervals are as follows. In \cref{fig:conv2_wo_bias_cali}, the dashed lines denote the representable range of 16-bit integers. In \cref{fig:conv2_w_bias_cali}, the dashed lines indicate the range of 8-bit integers. The portions that exceed the corresponding representable range are indicated by red bars.}
    \label{fig:bias_cali}
\end{figure*}
The original approach to quantization in ANNs for integer-only inference \cite{jacob2018quantization} involves the assumption that $S_b=S_w S_x$, which turns the \cref{eq:quantizeMAC} into 
\begin{align}
    \tilde{a}_i &= \hat{M} \sum_j (\tilde{W}_{ij}\tilde{X}_j + q_{b_i}), \notag \\
    \hat{M} &\equiv \frac{S_wS_x}{S_a}.
\end{align}
However, quantizing the bias based on this method typically requires higher precision. In resource-constrained hardware environments, it may be impossible to represent the full range of quantized bias values as required by the original algorithm, leading to failures in normal network inference.
For example, suppose the hardware supports a maximum storage bit-width of INT16, the activation values $a\in [-0.5,0.5]$ and weights $w \in [-0.2,0.2]$ are quantized to 8-bit integer, while the range of the bias is $[-2.5,2.5]$, according to \cref{quantize}, for $b=0.5\times 0.2$, the quantized value would be $2^{15}-1$.
However, in certain hardware-constrained environments (such as 16-bit or 8-bit), such bias will cause severe overflow, leading to distortion, as shown in \cref{fig:conv2_wo_bias_cali}. 
To reduce bits for storing the bias, we set $S_b=S_a$ instead. Then \cref{eq:quantizeMAC} can be simplified as:
\begin{align}
    \tilde{a}_i &= \max\{(\hat{M} \sum_j \tilde{W}_{ij}\tilde{X}_i) + q_{b_i},0\}, \label{eq:quantizeMAC_final} \\
    \hat{M} &\equiv \frac{S_wS_x}{S_b}\notag.
\end{align}
As shown in \cref{fig:conv2_w_bias_cali}, the distribution of quantized bias values is depicted when $S_b=S_a$. It is evident that an 8-bit integer can adequately represent the quantized bias in this scenario. 
Meanwhile, if a 16-bit integer environment for bias is available, we can decide whether to apply bias calibration by checking whether the quantized bias values significantly exceed the representable range of a 16-bit integer.

\subsection{STEM: From Quantized ANN to Hardware Compatible Model} \label{sec:Accumulating and Generating neuron model}
\paragraph{Accumulation: decode from spikes}

\begin{figure}[t]
    \centering
    \includegraphics[width=0.5\textwidth]{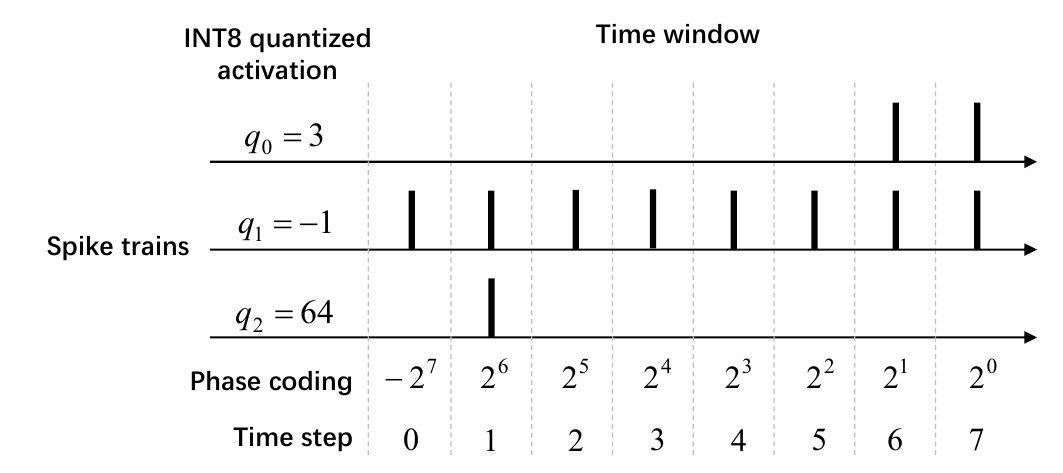}
    \caption{The spike train corresponding to the activation values after 8-bit quantization.}
    \label{fig:binary encoding}
\end{figure}
\begin{figure}[t]
    \centering
    \includegraphics[width=0.5\textwidth]{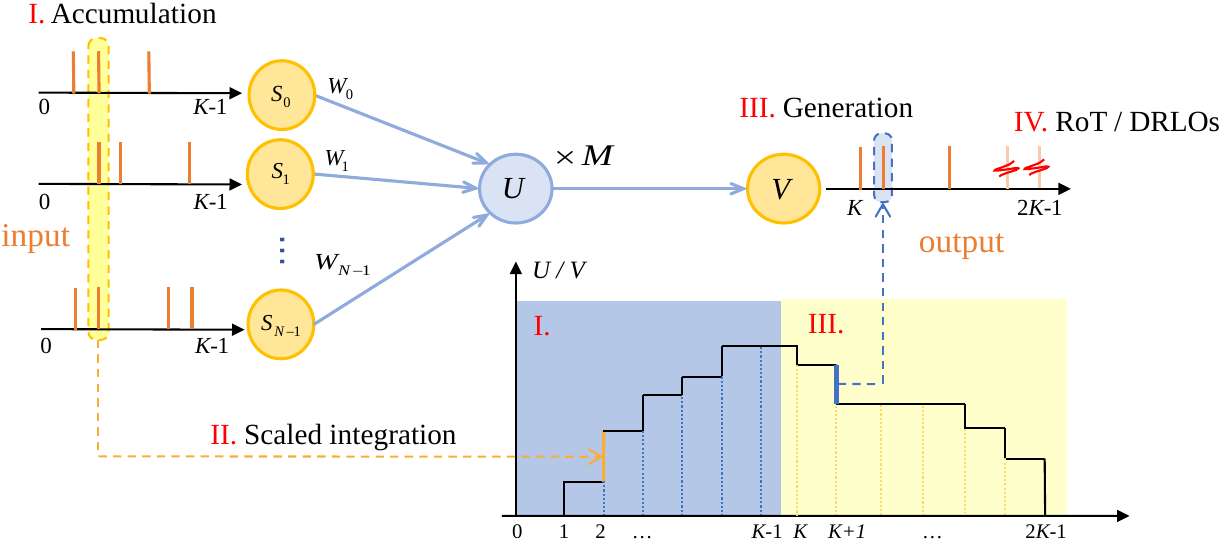}
    \caption{The working flow inside STEM model}
    \label{fig:multy}
\end{figure}
To perform the quantized ANN computation, we use a two-phase spiking encoding/decoding scheme, where the cell body is used to precisely compute the quantized input integers as well as output a spiking train that carries the quantized ANN output.
A signed integer activation $X_j$ from the previous layer, represented by $K$ bits, can be expressed as a power series in base $2$, as shown in \cref{eq:series} (an illustrative example is provided in \cref{fig:binary encoding}, with $K=8$).
\begin{align}
    \tilde{X}_j= -x_{j,K-1}\cdot2^{K-1} + \sum_{t=0}^{K-2} x_{j,t}\cdot2^t \label{eq:series}.
\end{align}
Note that all lowercase $x_i$ are in $\{0,1\}$. Now we use different time steps (indexed by $t$) to denote the different entries in the series, and we can get the binary form of a signed integer as a spike train. For example, as shown in \cref{fig:binary encoding}, the index of each time step $t$ here represents a specific binary bit.
For convenience, assume that the start time step is indexed $0$.
The decimal {\it $q=3$} can be rewritten as {\it 00000011} in binary form, then we use the spike trains $X_j=\{0,0,0,0,0,0,0,1,1\}$ in 8 time steps, i.e. $X_{j,0}=0,X_{j,1}=0,...,X_{j,6}=1,X_{j,7}=1$, to cast the integer $q$.
Then comes the four-phase scheme, as shown in \cref{fig:multy}.
In terms of series \cref{eq:series}, we can rewrite \cref{eq:MAC} as
\begin{align}
    I_i &= \sum_j \tilde{W}_{ij}\tilde{X}_j = \sum_j \tilde{W}_{ij} (-x_{j,K-1}\cdot2^{t-1} + \sum_{t=0}^{K-2} x_{j,t}\cdot2^t) \notag\\
    & = \sum_j \tilde{W}_{ij} \sum_{t=0}^{K-1} \Phi(t) \cdot x_{j,t}  = \sum_{t=0}^{K-1} \sum_j \Phi(t)\tilde{W}_{ij}\cdot x_{j,t}.
    \label{eq:spike_encoded_MAC}
\end{align}
$\Phi (t)$ is defined as 
\begin{align}
    \Phi (t) = \begin{cases}
        2^{K-t-1},    &0< t \leq K-1\\
        -2^{K-1},   &t=0
        \end{cases}. \label{eq:Phi}
\end{align}
Then we unroll the summation over time $t$, and then we can store the weighted sum at each time step $t$ :
\begin{align}
    I_{i,t} &= \sum_j \Phi(t)\tilde{W}_{ij} \cdot x_{j,t}. \label{update:I}
\end{align}

\paragraph{Scaled integration}
\begin{figure*}[ht]
    \centering
    \begin{subfigure}[b]{0.45\textwidth}
        \centering
        \includegraphics[width=\textwidth]{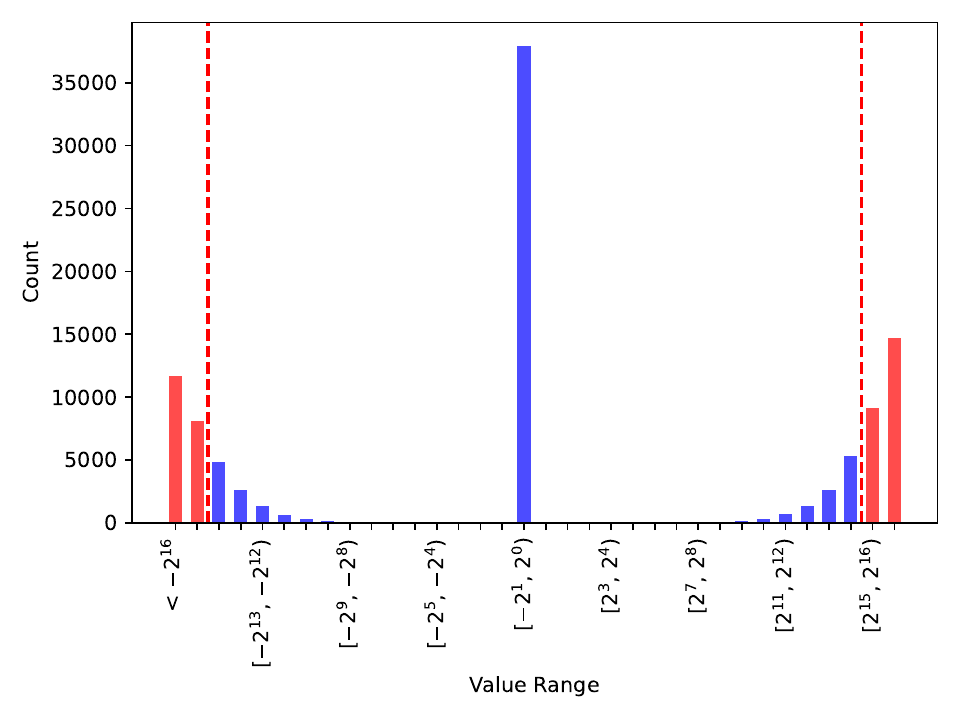}
        \caption{Without scaled integration.}
        \label{fig:fc_wo_scale}
    \end{subfigure}
    \hspace{1cm}
    \begin{subfigure}[b]{0.45\textwidth}
        \centering
        \includegraphics[width=\textwidth]{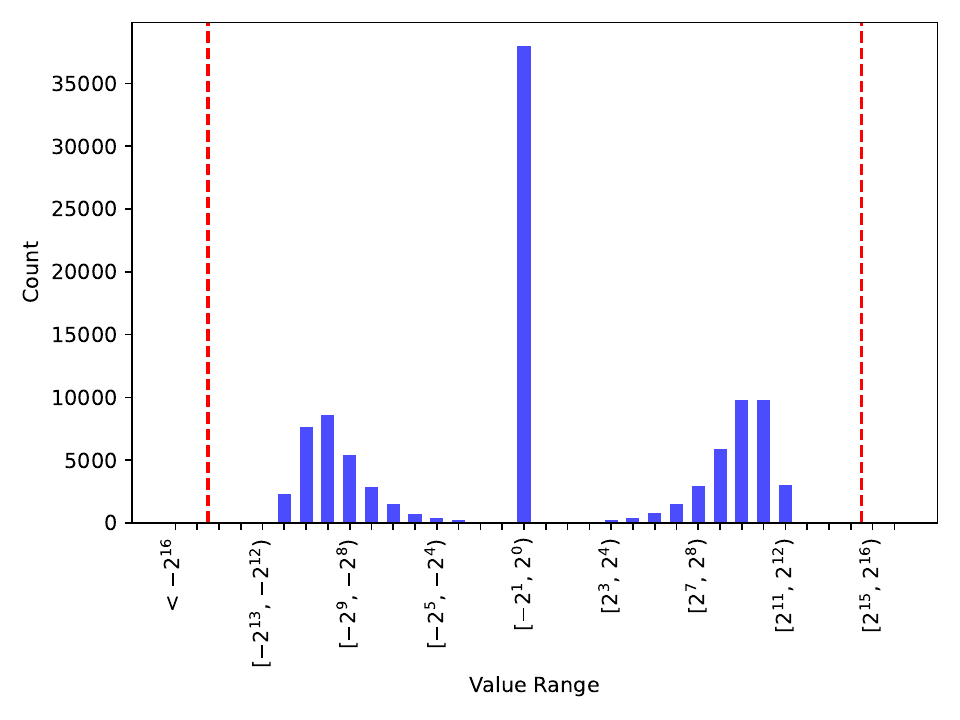}
        \caption{With scaled integration.}
        \label{fig:fc_w_scale}
    \end{subfigure}
    \caption{Distribution statistics of quantized synaptic current: The dashed lines indicate the representable range of 16-bit integer. The portions exceeding the corresponding representable range are highlighted with red bars.}
    \label{fig:safe_scale}
\end{figure*}
It should be noted that the total result, $I_{i}$ in \cref{eq:spike_encoded_MAC}, may overflow due to the constraint of finite hardware bit width.
Assuming that the receptive field of the neuron is $N$ , weight and activation are quantized to $K$ bits. The ideal range holds:
\begin{equation}
    -N\times2^{2K-1} \leq \sum_{j=0}^{N-1} \tilde{W}_{ij} \tilde{x}_j \leq N(2^{2K-1} - 1).
\end{equation}
However, this range likely exceeds the representational limit of the intermediate results.
For instance, when the intermediate result is represented by a 16-bit integer and choose $K=8$ for quantization purposes, $N=512$. Note that $2^{15} - 1 < 256 \times (2^{14} - 1)$. 
In order to guarantee that the results do not exceed the target integer capacity, additional bit width is necessary to store the intermediate results.
As demonstrated in \cref{fig:fc_wo_scale}, we extract the first fully-connected (FC) layer of Tiny-VGG and examine the distribution of synaptic current without the implementation of scaled integration. Statistics are from 100 images derived from ImageNet-1k.
Severe overflow is observed due to inadequate integer precision (targeting 16-bit), resulting in network failure (with accuracy on ImageNet below 2\%).
Hence, we propose the scaled integration to address this issue. This method has been demonstrated to effectively prevent overflow-induced information loss in the model.
First, we collect the maximum of $I_{\max}$ during ANN quantization. Scaled integration aims to scale $I_{\max}$ down so that it can be represented by an $n$-bit integer. Here we specify two scaling factors
\begin{align}
    M_0 &\equiv \frac{2^{n-1}-1}{I_{\max}},\\
    M_1 &\equiv \frac{\hat{M}}{M_0}.
\end{align}
During each weight summation calculation, the result is first scaled by multiplying it by $M_0$.
In view of the hardware, we can use a single hidden state to get the sum of $\sum_tI_{i,t}$ by
\begin{align}
    U_{i,t} &= U_{i,t-1} + M_0I_{i,t} \label{eq:The two-step scaling}
\end{align}
where the initial value is $U_{i,0}=0$. \cref{eq:The two-step scaling} can be executed by the neuromorphic cores where the variable $U_{i,t}$ is locally accumulated at time step $0\le t <K-1$.
After all spikes have been received, a second scaling is performed by multiplying the computed result by $M_1$ as
\begin{align}
    V_{i} &\equiv M_1U_{i,K-1}+\tilde{b_i}, \label{init:V_-1}
\end{align}
After applying scaled integration, the synaptic current distribution is scaled to a safe range, as shown in \cref{fig:fc_w_scale}. It is evident that $V_{i}= M_1M_0\sum_tI_{i,t}+\tilde{b_i}={\tilde M}\sum_j \tilde{W}_{ij}\tilde{X}_j+ \tilde{b_i}$.
So we get the weight sum within the ReLU part of \cref{eq:quantizeMAC_final} as well as \cref{eq:relu}.

\paragraph{Generation}
Next, we encode the activation values after the ReLU operation during the generation phase.
During this phase, the neuron acts as a
\begin{align}
    S_{i,t} &= \Theta (V_{i,t}-2^{2K-t-1}), \label{update:S_t} \\
    V_{i,t} &= V_{i,t-1}-2^{2K-t-1}S_{i,t}, \label{update:V_t}
\end{align}
where $\Theta$ is the Heaviside step function.
$S_t\in \{0,1\}$ represents the spike emitted at time step $t$ ($K\le t < 2K$).
At $t=K-1$, the final spike train is computed, and its corresponding binary value is assigned to $V_{K-1}$ according to \cref{init:V_-1}.
After a spike is emitted, the value corresponding to $V_t$ is subtracted.
Note that the $\Theta$ function ensures that $S_t$ is a non-negative value, where the negative value is clamped to zero. During the firing phase, to properly use the $\Theta$ function, we treat the membrane potential as an unsigned integer.
The multiplier of $2^{2K-t-1}$ ensures that the spike timing corresponds to the binary position of each bit.

\subsection{Reducing Spikes and Saving Energy}
\begin{figure}[ht]
    \centering
    \includegraphics[width=0.5\textwidth]{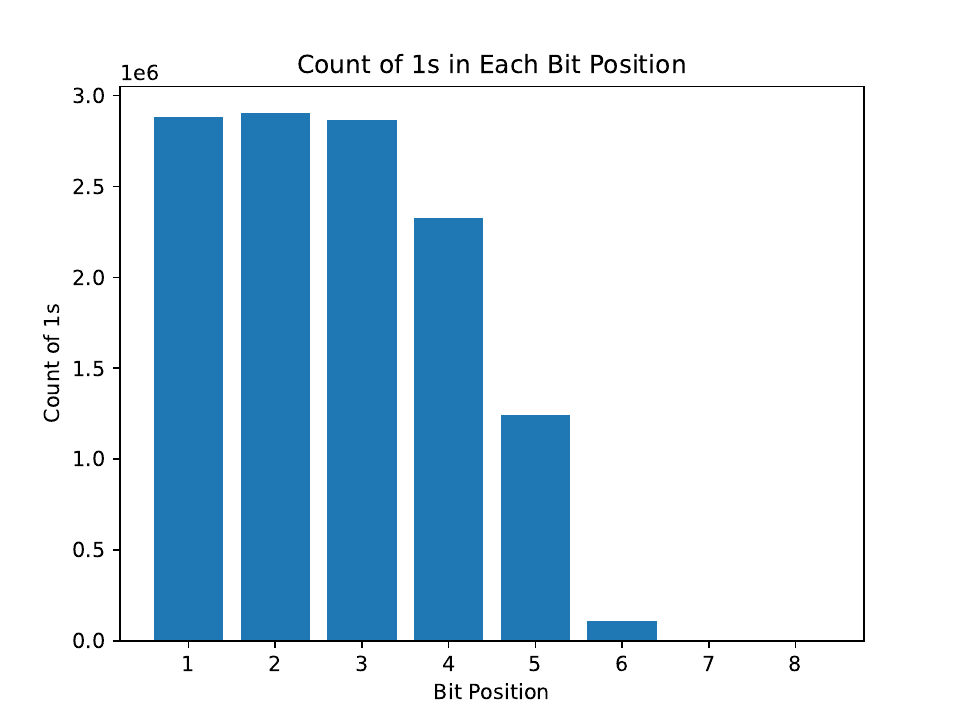}
    \caption{Statistics of the count of $1$ in each Bit position.}
    \label{fig:cnt1}
\end{figure}

In \cref{update:I}, $x_{j,t}$ values are confined to the set ${0,1}$ as the input spike train, where MACs are reduced to mere accumulation (ACs) operations.
When a spike arrives, the parameter $W$ is added to $I_t$ through a single addition operation, referred to as a Synaptic Operation (SOP).
Therefore, each spike in the sequence corresponds to one AC and one SOP. A reduction in spikes directly leads to reduced energy consumption.
As stated in \cite{horowitz20141}, the 8-bit integer MAC operation in a quantized ANN consumes 0.23 pJ (0.2 + 0.03 pJ), with the AC operation consuming only 0.03 pJ.
Meanwhile, substantial spikes can slow down the data transmission between the neuromorphic hardware and other peripheral devices.
Controlling the number of AC operations, as well as spikes over the network, can reduce the power consumption of neuromorphic hardware.
In the SDANN framework, spike trains are equivalent to mapping the binary activation values.
Therefore, a reduction in the number of {\it 1} in the activation value $V_{i,K-1}$ can effectively reduce spikes in the output. This, in turn, further reduce transfer spikes between layers inside the network.

As demonstrated in \cref{fig:cnt1}, we study the distribution of bits of the quantized activation value of the model. 
Data come from the result from the initial convolutional layer of the Tiny-VGG architecture with 100 images from the ImageNet-1k dataset.
The statistics indicate that the lower bits of the binary code carry less significant information compared to the higher bits.
The majority of the bits set to $1$ in the activation value distribution are located in the lower bits of the binary code. Reducing the number of $1$s in the lower bits leads to minimal information loss while significantly decreasing the number of SOPs.
In light of the findings, two methodologies are hereby proposed with the objective of mitigating the occurrence of $1$ in the quantized activation value.
It should be noted that the methods outlined in this subsection are \textbf{optional}. The basic functionality of the SDANN is not dependent on the methods in this subsection for reducing spikes.

\begin{figure*}[t]
    \centering
    \begin{tabular}{c c c}
        \begin{subfigure}[t]{0.45\textwidth}
            \centering
            \includegraphics[width=\linewidth]{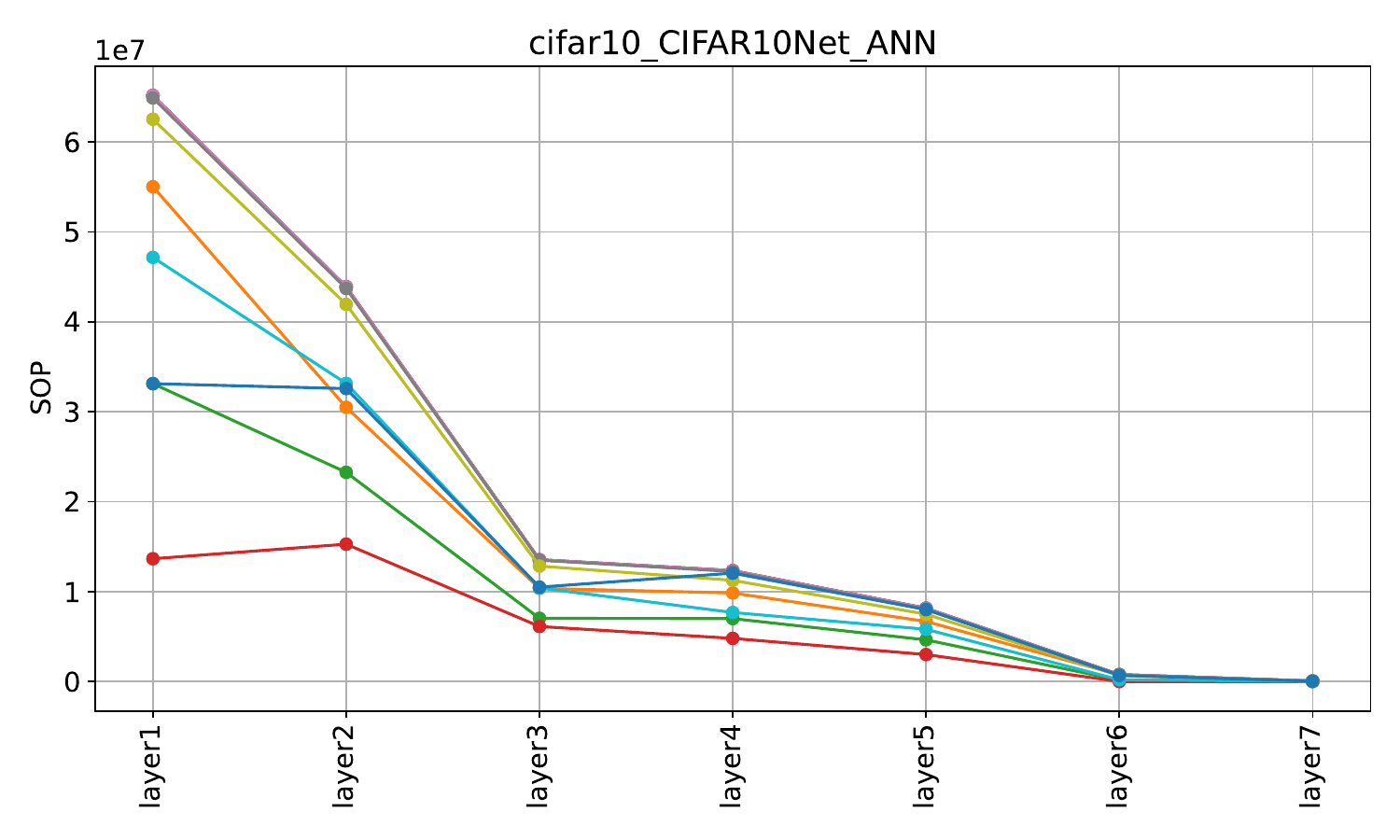}
            \caption{CIFAR10Net in CIFAR10}
        \end{subfigure}
        &
        \multirow{2}{*}{
            \includegraphics[width=0.1\textwidth]{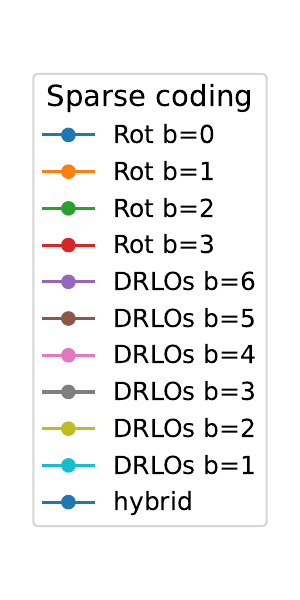}
        }
        &
        \begin{subfigure}[t]{0.45\textwidth}
            \centering
            \includegraphics[width=\linewidth]{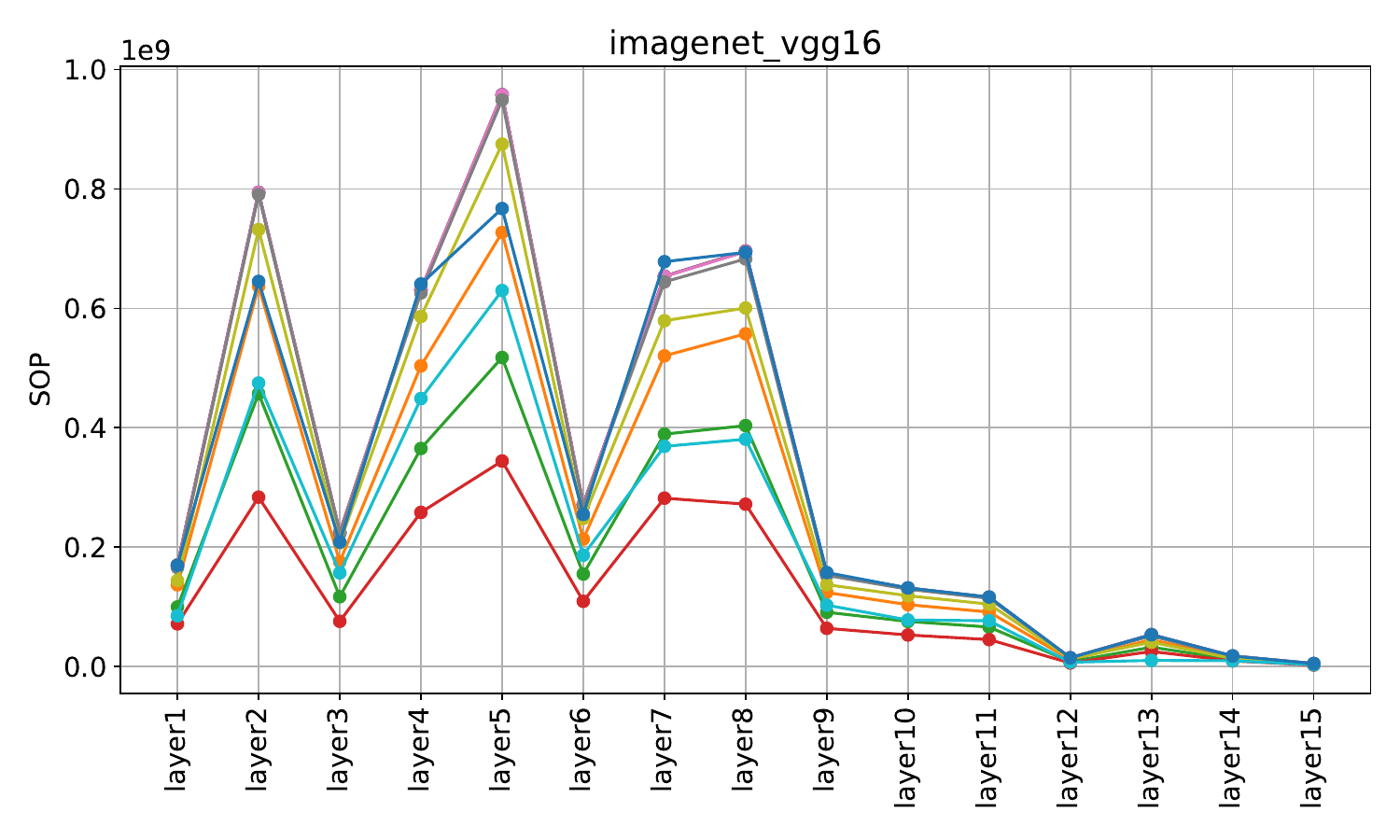}
            \caption{VGG-16 in ImageNet}
        \end{subfigure}
        \\
        \begin{subfigure}[t]{0.45\textwidth}
            \centering
            \includegraphics[width=\linewidth]{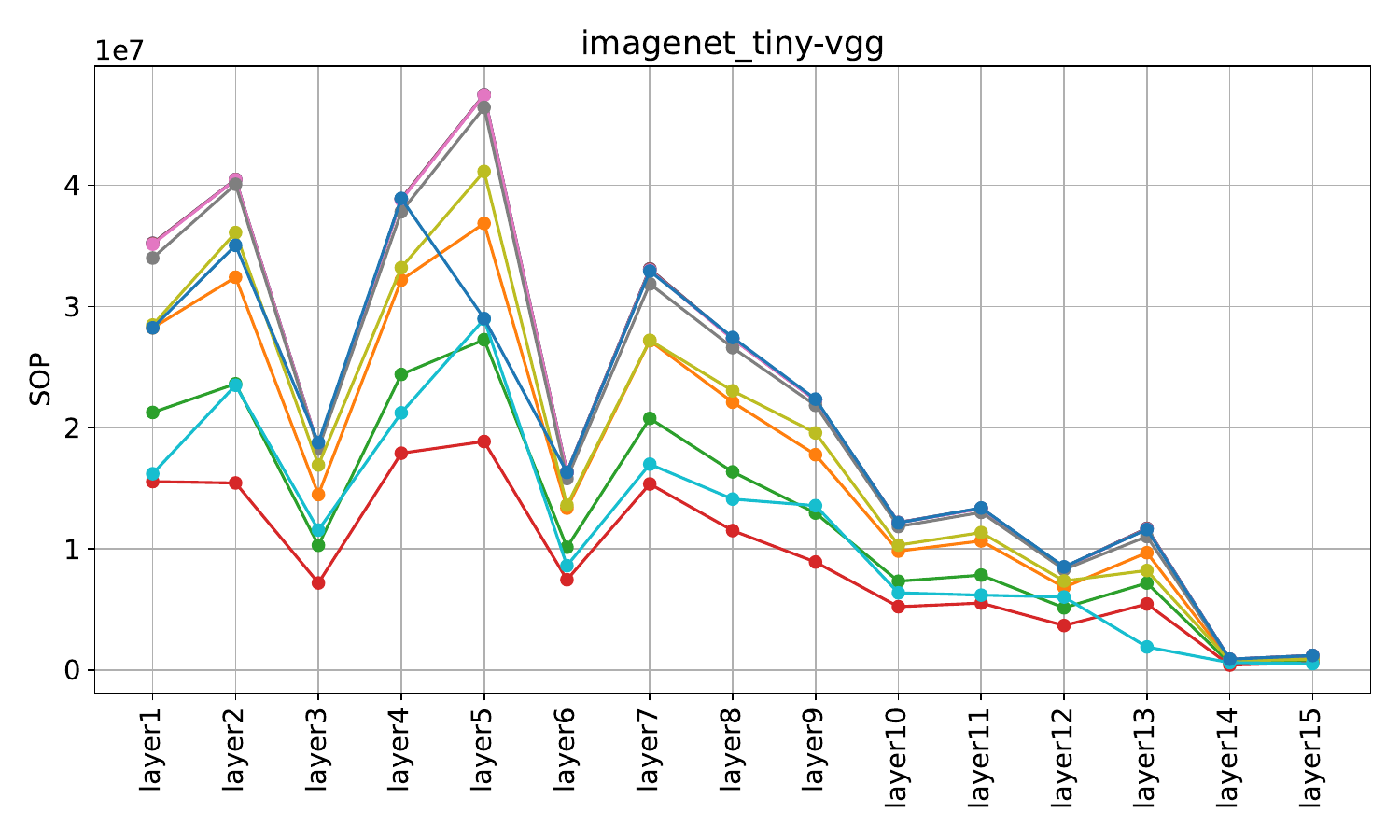}
            \caption{Tiny-VGG in ImageNet}
        \end{subfigure}
        &
        &
        \begin{subfigure}[t]{0.45\textwidth}
            \centering
            \includegraphics[width=\linewidth]{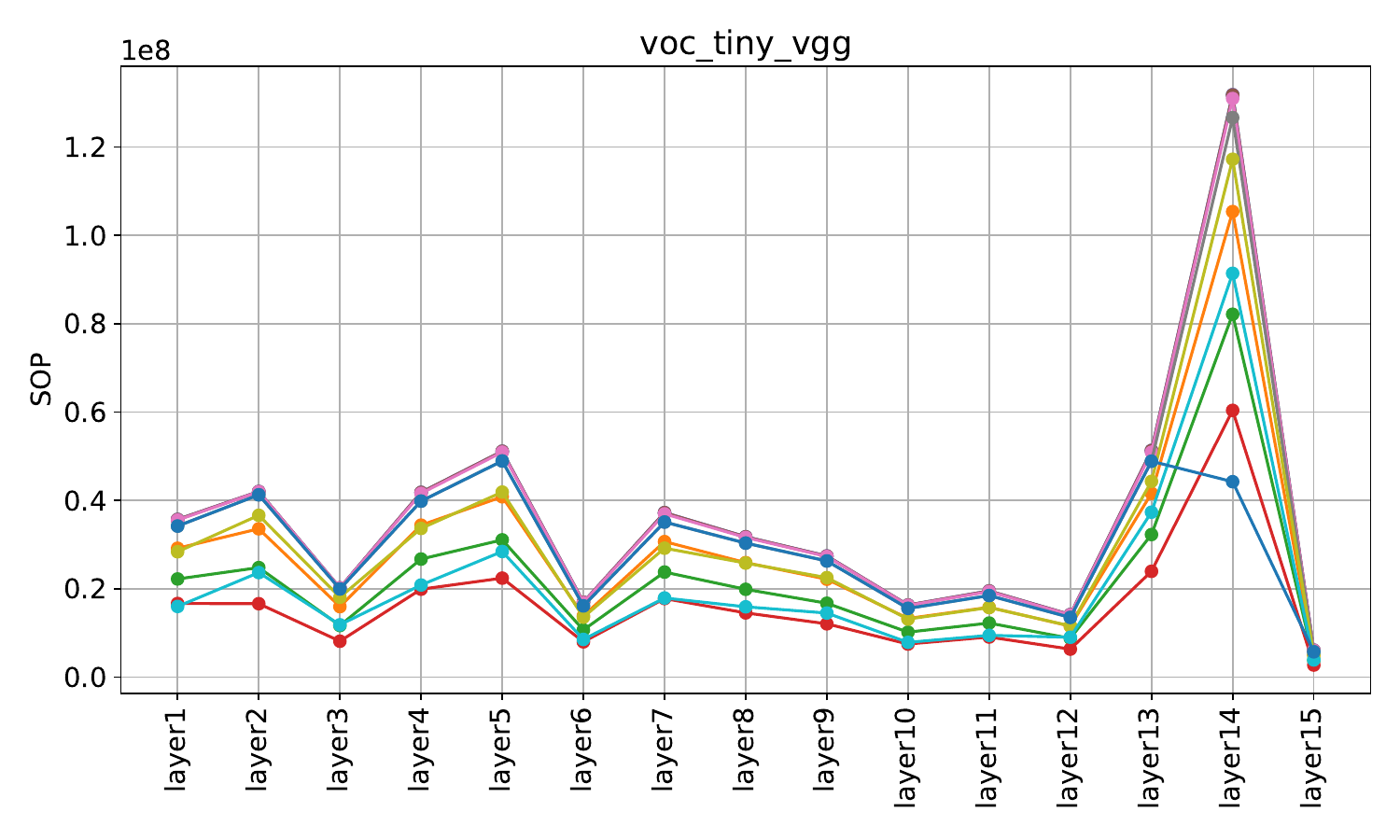}
            \caption{Tiny-VGG in VOC07}
        \end{subfigure}
    \end{tabular}
    \caption{The SOP of each layer under different spike sparsity schemes across various datasets and tasks.}
    \label{fig:spike_counts_layers}
\end{figure*}
\paragraph{Round-off Truncation (RoT)}
First, we count the SOPs across different layers of the model, as shown in \cref{fig:spike_counts_layers}. The number of spikes varies significantly between layers.
Therefore, we apply rounding and truncation to the activation values using layer-specific bit settings $\tilde K$, aiming to minimize spike counts (SOPs) while maintaining acceptable performance.
The rounded value can be expressed as
\begin{align}
    V_{i,K-1}' = round(\frac{V_{i,K-1}}{2^{b}})\cdot 2^{b}, \label{eq:Round off truncation}
\end{align}
where $V_{i,K-1}'$ are the activation values before and after spike sparsity, and $b$ is the target rounding bit width factor.

\paragraph{Discarding Redundant Low-Order spikes (DRLOs)} 
\begin{figure}[ht]
    \centering
    \includegraphics[width=0.5\textwidth]{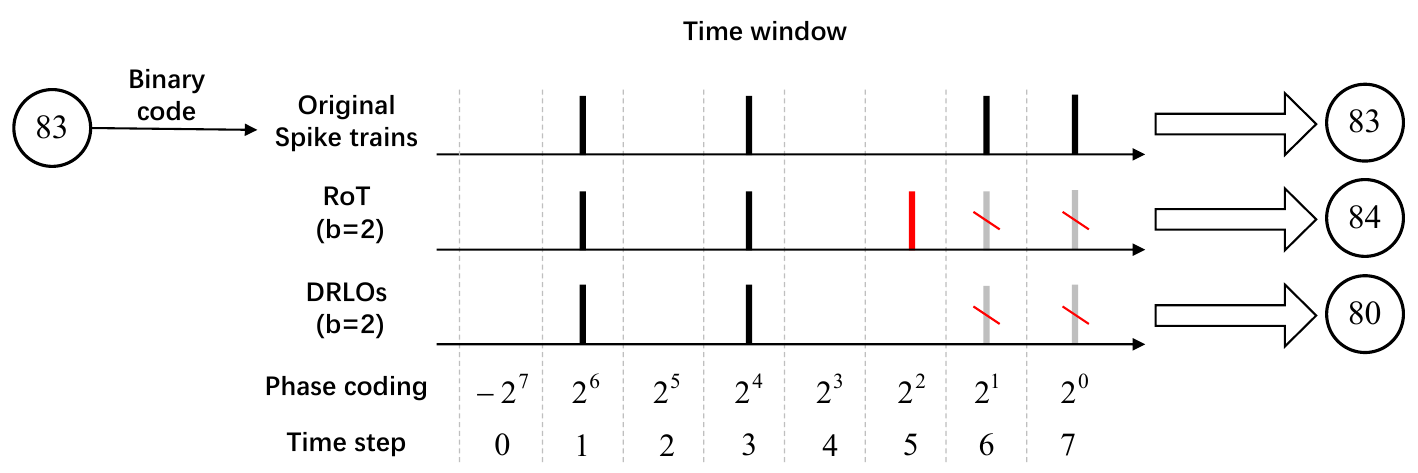}
    \caption{Spike sparsity schemes in 8-bit}
    \label{fig:round off and limit}
\end{figure}
As demonstrated in \cref{fig:cnt1}, results indicate that the information carried by higher-order bits is significantly greater than that of lower-order bits.
Hence, it is feasible to discard lower-order bits to further reduce spikes, with only minor performance degradation.
For quantized activation values, if the number of 1s in their corresponding binary code exceeds $b$, the lower-bit 1s are discarded so that the total number of 1s equals $b$. This ensures that the number of spikes emitted by a neuron during the generation phase does not exceed $b$.

As an example, \cref{fig:round off and limit} illustrates the sparsity of the spike trains obtained after 8-bit quantization using both methods.
For RoT, the spikes at time steps 6 and 7 are rounded to time step 5, changing the binary representation from $01010011$ to $01010100$.
For DRLOs, there are a total of $4$ spikes. To ensure that the total number of spikes is less than $2$, we discard the two spikes in the lower bits.

\subsection{Running in Pipeline}

\begin{figure*}[ht]
    \centering
    \includegraphics[width=\textwidth]{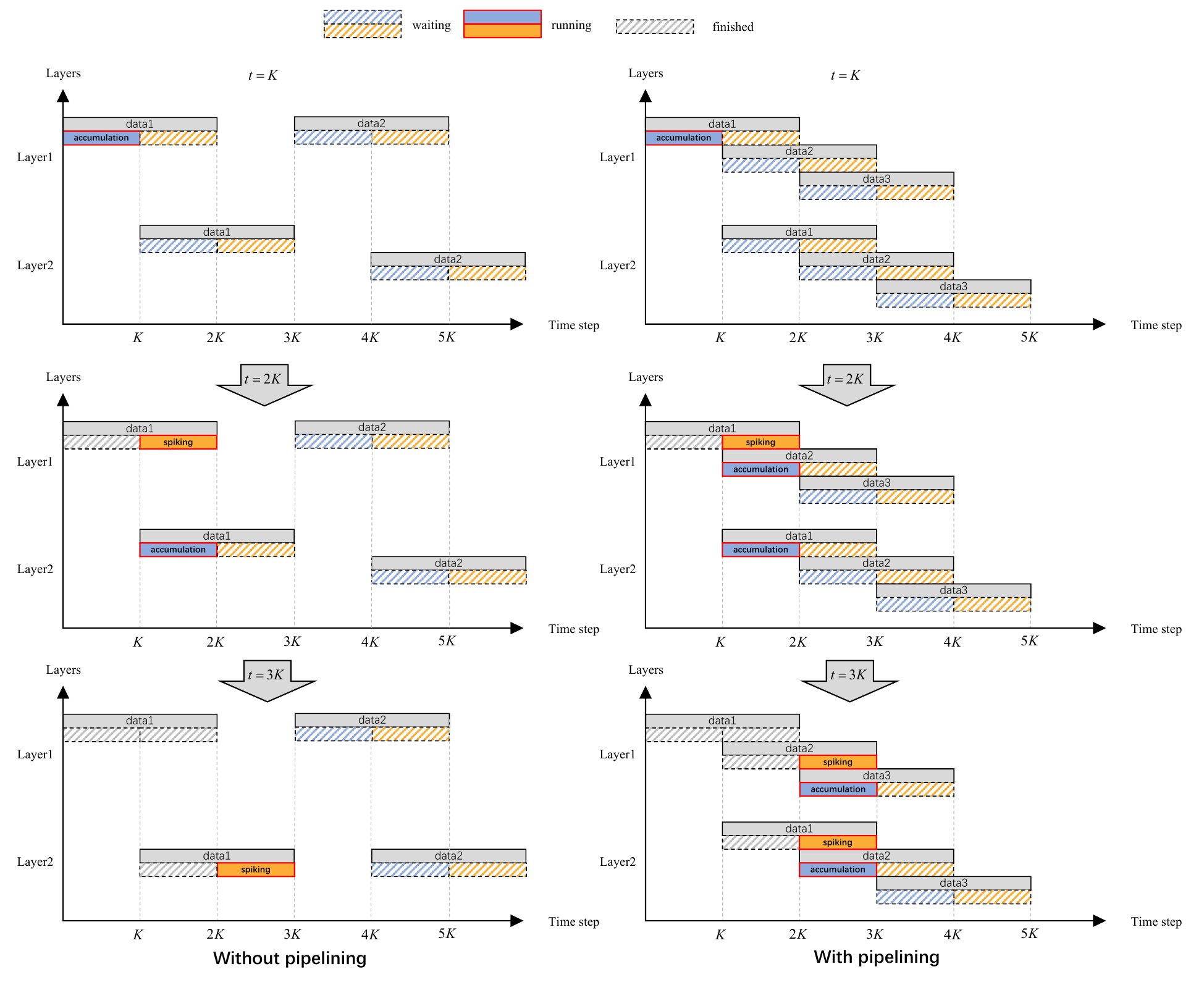}
    \caption{Pipelining of SDANN. For clarity, we take a simple 2-layer network as an example and define $K$ time steps as one processing cycle of STEM. The spike data requires a latency of $K$ time steps from reception to spike generation. By introducing a pipeline, different layers can process different inputs simultaneously, thereby reducing the average inference latency.}
    \label{fig:pipeline}
\end{figure*}

\begin{algorithm}[t]
    \caption{The process of STEM}
    \label{algorithm:STEM}
    \begin{algorithmic}[1] 
        \REQUIRE Input Number $N$, Quantization Bits $K$, RoT factor for each layer $\hat K$, DRLOs factor $\bar K$.
        \FOR{ neuron $i$ in the $l$-th layer}
            \STATE $U_{i,-1}=0$
            \STATE \# Accumulation \& scaled integration
            \FOR{ $t=0$ \textbf{to} $K-1$}
                \STATE update $U_{i,t}$ using \cref{update:I} and \cref{eq:The two-step scaling}
            \ENDFOR
            \STATE calculate $V_{i,K-1}$ by \cref{init:V_-1}
            \STATE do RoT or DRLOs (Optional)
            \STATE round off $V_{i,K-1}$ by \cref{eq:Round off truncation} with $\hat K$
            \STATE \# Generation:
            \FOR{ $t=K$ \textbf{to} $2K-1$}
                \STATE update $S_{i,t}$ by \cref{update:S_t}
                \STATE update $V_{i,t}$ by \cref{update:V_t}
            \ENDFOR
        \ENDFOR
    \end{algorithmic}
\end{algorithm}

During the accumulation and generation phases, each STEM neuron must wait for the completion of input accumulation before generating its output.
This scheme inevitably introduces redundant time-step delays during inference.
Following the approach in \cite{liu2022spikeconverter}, pipelining can be applied to improve throughput by allowing the network to process multiple inputs in parallel.
\cref{fig:pipeline} illustrates the operational stages of a two-layer example network.
As shown in \cref{update:I} to \cref{update:V_t}, each neuron requires $2K$ time steps to receive inputs and emit spikes.
Consequently, the entire model takes $2K$ time steps to process one input sample.
The operations of each layer can be overlapped such that the previous layer starts processing the next input while the subsequent layer is handling the current one.
As a result, the average inference time per sample in a fully pipelined setup is reduced to $K$, representing a significant improvement over the naive approach.

\section{Experiments and Analyses}\label{sec:experiments}
Before hardware deployment, we evaluate SDANN using software simulation based on the PyTorch framework \cite{pytorch}.
To evaluate the effectiveness of the proposed SDANN methodology, we conduct experiments on three tasks using different model architectures.
\begin{itemize}
\item Classification on CIFAR10: CIFAR100 \cite{cifar} has 50,000 images in 10 categories for training and 10000 for evaluation. Each image has a resolution of $32\times32$ with three color channels.
\item Classification on ImageNet-1k: ImageNet-1k is the training set and validation set of the ImageNet Large-Scale Visual Recognition Challenge in 2012 (ILSVRC2012) \cite{imagenet}. It contains over 1.2 million high-resolution images across 1,000 categories.
\item Object detection on VOC2007: PASCAL VOC 2007 (VOC2007) \cite{voc} datasets are a group of image datasets for different vision tasks. In this study, we focus on the object detection subset of the dataset.
\end{itemize}
These datasets have been extensively utilized for the evaluation of model performance both in the domains of ANNs and SNNs.

The SDANN models used in our experiments are based on the VGG and ResNet architectures.
VGG represents a typical convolutional ANN without shortcut connections.
ResNet employs the residual path as the fundamental component of modern ANNs.
For the task of object detection, we utilize YOLOv1-based network\cite{yolo}.
Furthermore, based on hardware constraints and neuron count limits, we scale down the original architectures to construct two smaller models, namely Tiny-VGG and Tiny-YOLO. Tiny-YOLO adopts Tiny-VGG as its backbone.

All experiments are conducted using PyTorch \cite{pytorch} on a workstation equipped with eight NVIDIA GeForce RTX 3080 GPUs, each with 10GB memory, CUDA version 12.2, and driver version 535.230.02. We use stochastic gradient descent (SGD) with a momentum of 0.9 and weight decay of $5 \times 10^{-4}$. The initial learning rate is 0.1 for image classification tasks and 0.001 for object detection.

For classification on ImageNet using VGG16, ResNet-18, and ResNet-34, the input image size is $224\times224$. For object detection on VOC using the corresponding pre-trained backbones, the input size is $448\times448$. For Tiny-VGG and Tiny-YOLO, the input size is $128\times128$. We report Top-1 accuracy for classification and mAP@0.5 for object detection to compare model performance.

\subsection{Performance of SDANN on various tasks}\label{subsec:Overall Performance}

\Cref{tab:Overall Performance} presents the evaluation of our models with different architectures, considering metrics such as performance accuracy, weight precision, conversion loss, and required time steps across various datasets and tasks.
In these experiments, RoT and DRLOs are not applied for spike sparsity.
\textbf{Notably, the performance of the SDANN model is equivalent to that of the quantized ANN model.}
This alignment is due to the direct mapping of SDANN with STEM to the corresponding quantized ANN.

\begin{table*}[ht]
\centering
\caption{Performance of SDANN}
\label{tab:Overall Performance}
\begin{tabular}{ccccccccc}
        \toprule
        ~                                    & Architecture & \makecell{Precision\\(ANN W/A)} & \makecell{\textbf{ANN}\\ \textbf{Acc.}}&\makecell{Precision\\(SDANN W)} & \makecell{\textbf{SDANN}\\ \textbf{Acc.}} &  \makecell{\textbf{SDANN - ANN}\\ \textbf{Acc. (\%)}} & \makecell{\textbf{Time}\\ \textbf{Steps}}\\
        \midrule
        
        \multirow{2}{*}{{\makecell{\textbf{CIFAR10}}}}
        ~                               & VGG-16    & 8/8   & 92.87 & \textbf{8}  & 92.87 & \textbf{0}  & 8 \\
        ~                               & ResNet-18 & 8/8   & 93.07 & \textbf{8}  & 93.07 & \textbf{0}  & 8 \\
        
        \midrule
        \multirow{4}{*}{{\makecell{\textbf{ImageNet}}}} 
        ~                               & Tiny-VGG  & 8/8   & 53.60 & \textbf{8}  & 53.60 & \textbf{0}  & 8 \\
        ~                               & VGG-16    & 8/8   & 68.28 & \textbf{8}  & 68.28 & \textbf{0}  & 8 \\
        ~                               & ResNet-18 & 8/8   & 68.02 & \textbf{8}  & 68.02 & \textbf{0}  & 8 \\
        ~                               & ResNet-34 & 8/8   & 70.16 & \textbf{8}  & 70.16 & \textbf{0}  & 8 \\
        \midrule
        \multirow{2}{*}{\makecell{\textbf{VOC2007}\\ \textbf{mAP}}} 
                 & Yolov1(Tiny-VGG)    & 8/8 &  50.51 & 8 & 50.51 & \textbf{0}  & 8 \\
                 & Yolov1(ResNet-34)   & 8/8 &  72.66 & 8 & 72.66 & \textbf{0}  & 8 \\
        \bottomrule
    \end{tabular}
\end{table*}
\subsection{Ablation study}
  
\subsubsection{Comparison with the Direct Training Method in CIFAR10}
\begin{figure}[ht]
\centering
\begin{subfigure}[t]{0.45\textwidth}
    \centering
    \includegraphics[width=\textwidth]{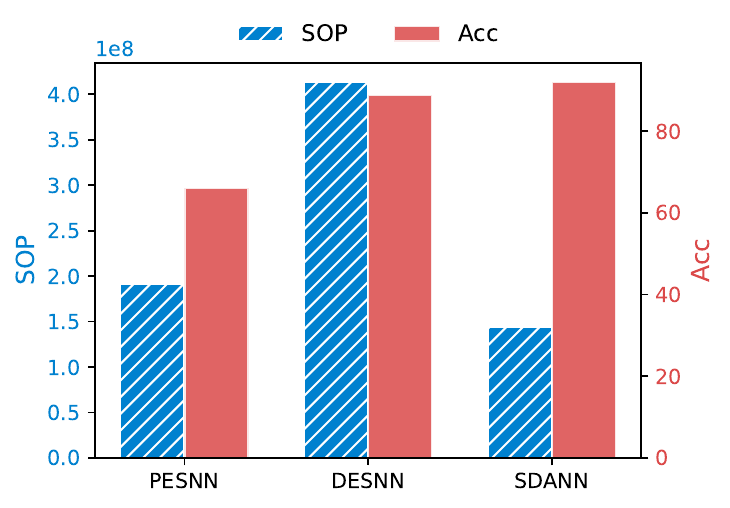}
    \caption{SOP and accuracy of models with different encoding methods.}
    \label{fig:cifar10-snn-vs-ours}
\end{subfigure}
\begin{subfigure}[t]{0.45\textwidth}
    \centering
    \includegraphics[width=\textwidth]{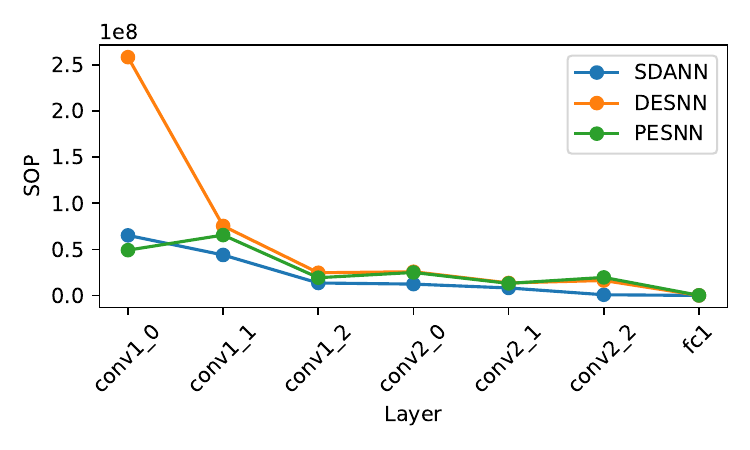}
    \caption{SOPs for each layer of models with different encoding methods.}
    \label{fig:cifar10-spike-counts}
\end{subfigure}
\caption{Comparison with directly trained SNN methods using different encoding schemes on CIFAR-10. Here, PEDSNN and DESNN represent SNNs with Poisson encoded SNN and directly encoded SNN, respectively.}
\label{fig:cifar10-subfigures}
\end{figure}

In the context of spike trains, the STEM approach in our SDANN can be regarded as a novel method for encoding spike timing information.
To evaluate its effectiveness, we conducted experiments comparing SDANN with two widely utilized SNN encoding methodologies: direct encoding, as used in \cite{wu2018stbp}, and position encoding as adopted in \cite{xu_qi_cvpr_kdsnn}.
The CIFAR10Net structure was adopted from \cite{fang2021plif} for the CIFAR10 classification method, with LIF neurons utilized for the SNN.
In contrast, the base ANN of SDANN replaces LIF neurons with ReLU layers. Additionally, a one-dimensional average pooling layer is used in place of the original voting layer in CIFAR10Net.
The quantization bit width is set to 8, and the corresponding time window for the SNN is also set to 8.
The SOP statistics are derived from SDANN models during inference per image from the CIFAR10 test set, excluding the input and output layers.
\cref{fig:cifar10-spike-counts} presents the total SOP, along with the accuracy achieved on the test set.
As shown, our method produces significantly sparser spikes in comparison to frequency-based encoding methods.
This finding suggests that the spiking model of SDANN can attain sparsity comparable to that of conventional SNNs.
Additionally, SOPs for each layer of the SNNs and the model under consideration are documented in \cref{fig:cifar10-snn-vs-ours}.
As the depth of the network is increased, the number of spikes experiences a gradual decrease.
Our method generates fewer spikes, particularly in the early layers.

\subsubsection{The Impact of scaled integration}

\begin{table*}[h]
    \centering
    \caption{The impact of quantization and scaling}
    \label{The impact of quantization and scaling}
    \begin{tabular}{ccccc}
        \toprule
        ~                                      & \multirow{2}{*}{Model Architecture}              & Full-precision ANN & Quantized ANN & SDANN\\
        ~                                      &               & - & + Quantized & + Quanized + scaled integration\\
        \midrule
        \multirow{2}{*}{CIFAR10(Acc.\%)}  & CIFAR10Net & 92.29                & 91.93(-0.36)             & 91.89(-0.40)  \\
        & ResNet-18 & 93.26                & 93.09(-0.17)             & 93.01(-0.25)  \\
        \midrule
        \multirow{4}{*}{ImageNet(Acc.\%)} & VGG-16                         & 71.04                & 70.03(-1.01)       & 69.97(-1.07)  \\
        ~                                      & ResNet-18                      & 70.48                & 69.13(-1.35)       & 69.08(-1.40)  \\
        ~                                      & ResNet-34                      & 73.19                & 71.73(-1.46)       & 71.64(-1.55)  \\
        ~                                      & Tiny-VGG                       & 55.56                & 53.65(-1.91)       & 53.60(-1.96)  \\
        \midrule
        \multirow{2}{*}{VOC2007(mAP.\%)}  & Tiny-Yolo                     & 50.49                & 50.41(-0.08)        & 50.51(+0.02)  \\
        ~                                      & ResNet-34                     & 73.63                & 72.82        & 72.66(-0.16)  \\
        \bottomrule
    \end{tabular}
\end{table*}

\Cref{The impact of quantization and scaling} reports the performance of full-precision ANNs, quantized ANNs, and SDANNs with and without scaled integration across various tasks.
It should be noted that if the hardware has sufficient storage to retain the accumulated weighted sum $U_{i,t}$ in \cref{eq:The two-step scaling}, this scaling process can be omitted.
Without scaled integration, the SDANN exhibits the same behavior, mirroring the quantized ANN, and displays equivalent performance as evidenced \cref{tab:Overall Performance}.
However, in the hardware utilized in this study, such as the Darwin3, scaled integration is imperative, given that each computing core is only capable of accommodating the 16-bit weighted sum value when the synaptic weight is configured to 8-bit.
Nevertheless, in comparison with the accuracy of the quantized model, the incorporation of scaled integration exerts a negligible influence on accuracy. And in some cases, it even slightly increases the performance a little bit.
These results indicate that scaled integration does not significantly affect model performance.

\subsubsection{Spike Sparsification Method}

\begin{table*}[ht]
\centering
\caption{Impact of different values of $b$ in RoT methods on model performance.}
\begin{tabular}{lccccc}
\toprule
\multirow{2}{*}{Dataset} & \multirow{2}{*}{Model} & \multirow{2}{*}{Original Acc/mAP(\%)} & \multicolumn{3}{c}{Acc/mAP($\Delta$Acc/mAP)} \\
\cmidrule(lr){4-6}
        &       &          & $b=1$  & $b=2$  & $b=3$  \\
\midrule
\multirow{3}{*}{CIFAR-10} 
  &  CIFAR10Net  & 91.89 & 90.81(-1.08) & 86.68(-5.21) & 38.52(-53.37)\\
  &  ResNet18  & 93.01 & 92.55(-0.46) & 92.01(-1.00) & 89.15(-3.86)\\
  &  VGG16  & 92.91 & 92.47(-0.44) & 92.32(-0.59) & 90.40(-2.51)\\
\midrule
\multirow{4}{*}{ImageNet} 
  &  ResNet-18  & 69.08 & 65.67(-3.40) & 62.75(-6.32) & 40.56(-28.52)\\
  &  ResNet-34  & 71.64 & 57.50(-14.13) & 50.64(-21.00) & 19.78(-51.86)\\
  &  VGG16  & 69.97 & 67.64(-2.33) & 64.33(-5.64) & 41.98(-27.99)\\
  &  Tiny-VGG  & 53.60 & 53.52(-0.07) & 53.17(-0.43) & 41.80(-11.80)\\
\midrule
\multirow{2}{*}{VOC07}
 &  Yolov1(ResNet-34)  & 72.66 & 70.14(-2.52) & 68.89(-3.77) & 58.14(-14.52)\\
 &  Yolov1(Tiny-VGG)  & 50.51 & 49.42(-1.08) & 48.89(-1.62) & 44.95(-5.56)\\
\bottomrule
\end{tabular}

\label{tab:rot_acc}
\end{table*}

\begin{table*}[ht]
\centering
\caption{Impact of different values of $b$ in RoT methods on SOPs of models.}
\begin{tabular}{lccccc}
\toprule
\multirow{2}{*}{Dataset} & \multirow{2}{*}{Model} & \multirow{2}{*}{Original SOPs} & \multicolumn{3}{c}{SOPs(SOP reduction)} \\
\cmidrule(lr){4-6}
        &       &          & $b=1$ & $b=2$ & $b=3$ \\
\midrule
\multirow{3}{*}{CIFAR-10} 
 &  CIFAR10Net  & \num{1.44e+08} & \num{1.13e+08}(-21.31\%) & \num{7.53e+07}(-47.69\%) & \num{4.29e+07}(-70.18\%)\\
 &  ResNet18  & \num{5.92e+07} & \num{4.78e+07}(-19.30\%) & \num{3.45e+07}(-41.76\%) & \num{2.34e+07}(-60.58\%)\\
 &  VGG16  & \num{9.91e+07} & \num{8.08e+07}(-18.52\%) & \num{5.98e+07}(-39.72\%) & \num{4.17e+07}(-57.94\%)\\
\midrule
\multirow{4}{*}{ImageNet} 
 &  ResNet18  & \num{1.19e+09} & \num{9.57e+08}(-19.45\%) & \num{7.00e+08}(-41.10\%) & \num{4.86e+08}(-59.04\%)\\
 &  ResNet34  & \num{2.36e+09} & \num{1.96e+09}(-16.90\%) & \num{1.43e+09}(-39.39\%) & \num{1.02e+09}(-56.90\%)\\
 &  VGG16  & \num{4.89e+09} & \num{3.86e+09}(-20.99\%) & \num{2.79e+09}(-42.92\%) & \num{1.90e+09}(-61.13\%)\\
 &  Tiny-VGG  & \num{2.78e+08} & \num{2.22e+08}(-20.21\%) & \num{1.63e+08}(-41.38\%) & \num{1.15e+08}(-58.58\%)\\
\midrule
\multirow{2}{*}{VOC07} 
&  Yolov1(ResNet34)  & \num{1.09e+10} & \num{8.97e+09}(-17.80\%) & \num{6.66e+09}(-38.98\%) & \num{4.79e+09}(-56.14\%)\\
 &  Tiny-Yolo  & \num{5.25e+08} & \num{4.22e+08}(-19.57\%) & \num{3.22e+08}(-38.54\%) & \num{2.34e+08}(-55.37\%)\\
\bottomrule
\end{tabular}
\label{tab:rot_sop}
\end{table*}

\begin{table*}[ht]
\centering
\caption{Impact of different values of $b$ in DRLO methods on model performance.}
\begin{tabular}{llcccc}
\toprule
\multirow{2}{*}{Dataset} & \multirow{2}{*}{Model} & \multirow{2}{*}{Original Acc/mAP(\%)} & \multicolumn{3}{c}{Acc/mAP($\Delta$Acc/mAP)} \\
\cmidrule(lr){4-6}
        &       &          & $b=4$  & $b=3$  & $b=2$  \\
\midrule
\multirow{3}{*}{CIFAR-10} 
 &  CIFAR10Net  & 91.89 & 91.81(-0.08) & 91.90(0.01) & 91.40(-0.49)\\
 &  ResNet-18  & 93.01 & 93.11(0.10) & 92.89(-0.12) & 90.41(-2.60)\\
 &  VGG16  & 92.91 & 92.81(-0.10) & 92.89(-0.02) & 92.12(-0.79)\\
\midrule
\multirow{4}{*}{ImageNet} 
 &  ResNet-18  & 69.08 & 69.06(-0.02) & 68.77(-0.30) & 59.49(-9.59)\\
 &  ResNet-34  & 71.64 & 71.66(0.02) & 70.88(-0.76) & 45.96(-25.67)\\
 &  VGG16  & 69.97 & 69.99(0.02) & 69.96(-0.01) & 64.46(-5.51)\\
 &  Tiny-VGG  & 53.60 & 53.52(-0.08) & 53.17(-0.43) & 41.80(-11.80)\\
\midrule
\multirow{2}{*}{VOC07}
&  Yolov1(ResNet-34)  & 72.66 & 73.04(0.38) & 73.52(0.86) & 65.26(-7.40)\\
&  Yolov1(Tiny-VGG)  & 50.51 & 50.41(-0.10) & 50.20(-0.31) & 43.57(-6.94)\\
\bottomrule
\end{tabular}
\label{tab:drlo_acc}
\end{table*}

\begin{table*}[ht]
\centering
\caption{Impact of different values of $b$ in DRLO methods on SOPs of models.}
\begin{tabular}{llcccc}
\toprule
\multirow{2}{*}{Dataset} & \multirow{2}{*}{Model} & \multirow{2}{*}{Original SOPs} & \multicolumn{3}{c}{SOPs(SOP reduction)} \\
\cmidrule(lr){4-6}
        &       &          & $b=4$ & $b=3$ & $b=2$ \\
\midrule
\multirow{3}{*}{CIFAR-10} 
 &  CIFAR10Net  & \num{1.44e+08} & \num{1.44e+08}(-0.03\%) & \num{1.43e+08}(-0.49\%) & \num{1.37e+08}(-4.95\%)\\
 &  ResNet-18  & \num{5.92e+07} & \num{5.92e+07}(-0.13\%) & \num{5.80e+07}(-2.12\%) & \num{5.22e+07}(-11.98\%)\\
 &  VGG16  & \num{9.91e+07} & \num{9.86e+07}(-0.57\%) & \num{9.52e+07}(-3.94\%) & \num{8.30e+07}(-16.24\%)\\
\midrule
\multirow{4}{*}{ImageNet} 
 &  ResNet-18  & \num{1.19e+09} & \num{1.19e+09}(-0.10\%) & \num{1.17e+09}(-1.74\%) & \num{1.05e+09}(-11.70\%)\\
 &  ResNet-34  & \num{2.36e+09} & \num{2.36e+09}(-0.13\%) & \num{2.31e+09}(-2.21\%) & \num{2.05e+09}(-13.07\%)\\
 &  VGG16  & \num{4.89e+09} & \num{4.89e+09}(-0.06\%) & \num{4.83e+09}(-1.22\%) & \num{4.40e+09}(-9.98\%)\\
 &  Tiny-VGG  & \num{2.78e+08} & \num{2.77e+08}(-0.07\%) & \num{2.74e+08}(-1.47\%) & \num{2.48e+08}(-10.70\%)\\
\midrule
\multirow{2}{*}{VOC07} 
 &  Yolov1(ResNet34)  & \num{1.09e+10} & \num{1.09e+10}(-0.23\%) & \num{1.06e+10}(-2.68\%) & \num{9.44e+09}(-13.53\%)\\
 &  Tiny-Yolo  & \num{5.25e+08} & \num{5.23e+08}(-0.42\%) & \num{5.06e+08}(-3.66\%) & \num{4.47e+08}(-14.83\%)\\
\bottomrule
\end{tabular}
\label{tab:drlo_sop}
\end{table*}

\begin{table*}[t] 
\centering 
\caption{Comparison between the spike sparsity-enhanced model and the original model in terms of accuracy and SOP.} 
\label{tab:sparse coding} 

\begin{tabular}{@{}cccccc@{}}
    \toprule
    \multirow{2}{*}{\textbf{Dataset}} & \multirow{2}{*}{\textbf{Network}} & \multicolumn{2}{c}{\textbf{without spike sparsity}} & \multicolumn{2}{c}{\textbf{with spike sparsity (hybrid scheme)}} \\
    \cmidrule(lr){3-6}
                                       &            & Acc or mAP   & SOPs       & Acc or mAP   & SOPs   \\
    \midrule
    \multirow{2}{*}{\textbf{CIFAR10}}  & CIFAR10Net & 91.89 & \num{1.44e+08} & 90.86(-1.03) & \num{9.70e+07}(-32.58\%)\\
    & ResNet-18 & 93.01 & \num{5.92e+07} & 92.77(-0.24) & \num{5.54e+07}(-6.44\%)\\
    \midrule
    \multirow{4}{*}{\textbf{ImageNet}} & VGG-16     & 69.97 & \num{4.89e+09} & 69.17(-0.80) & \num{4.55e+09}(-6.92\%)  \\
                                       & Tiny-VGG   & 54.05 & \num{3.27e+08} & 53.37(-0.68) & \num{2.97e+08}(-9.32\%)  \\
                                       & ResNet-18  & 69.08 & \num{1.19e+09} & 68.18(-0.90) & \num{1.13e+09}(-4.88\%)  \\
                                       & ResNet-34  & 71.64 & \num{2.36e+09} & 70.26(-1.37) & \num{2.30e+09}(-2.65\%)  \\
    \midrule
    \multirow{2}{*}{\textbf{VOC2007}} & ResNet-34     & 72.66 & \num{1.09e+10} & 73.16(+0.49) & \num{1.02e+10}(-6.55\%)  \\
    & Tiny-Yolo    & 50.51 & \num{5.25e+08} & 49.96(-0.55) & \num{4.28e+08}(-18.58\%)\\
    \bottomrule
\end{tabular}
\end{table*}
In this study, we evaluate the impact of two optional spike sparsification methods, RoT and DRLO, on SOPs and performance. 
The same sparsification strategy is applied to the output of each layer in the model and evaluated the impact of different $b$ values for RoT and DRLOs on overall model performance and average SOP per sample, as shown in \cref{tab:rot_acc,tab:rot_sop,tab:drlo_acc,tab:drlo_sop}.
It is evident that an increase in the RoT factor leads to a corresponding decrease in the SOP sum.
This is because RoT rounds off more bits, thereby reducing the number of output spikes.
For DRLOs, it can be observed in \cref{tab:drlo_acc,tab:drlo_sop} that when $b>2$, the effect on both model accuracy and SOP count is minimal. This indicates that the majority of the output spike sequences in the model are inherently sparse.

The goal of these sparsification techniques is to reduce the number of spikes while preserving as much of the original accuracy as possible.
In practice, we adopt a hybrid strategy that combines the RoT and DRLOs. We determine whether to apply the spike sparsification method for each layer and select different sparsification strategies and corresponding parameters for the layers that require it. The overall impact on the model is summarized in \cref{tab:sparse coding}.
As results show, the hybrid one has the capacity to maintain accuracy while simultaneously reducing the total sum of SOP.

Furthermore, we estimate the theoretical energy consumption of the SDANN model in comparison to the quantized ANN.
This estimation adopts the criterion in \cite{horowitz20141}, which provides energy cost per operation in 45nm technology. Specifically, we use the reported values for 8-bit integer operations: an 8-bit AC consumes 0.03 pJ, and an 8-bit MAC operation consumes 0.23 pJ in total, which is comprising of 0.2 pJ for multiplication and 0.03 pJ for addition. Based on these values, we convert the total number of MACs in the ANN and the number of additions in the SDANN into estimated energy costs. The energy consumption is averaged over a single input sample. The results in \cref{tab:evaluation_energy} demonstrate that SDANN achieves significantly lower energy cost compared to the quantized ANN, highlighting its superior efficiency.
\begin{table*}[h] 
    \centering 
    \caption{Energy Consumption Estimation of Quantized ANN and SDANN.} 
    \label{tab:evaluation_energy} 

    \begin{tabular}{@{}cccccc@{}}
        \toprule
        \multirow{2}{*}{\textbf{Dataset}}  & \multirow{2}{*}{\textbf{Network}} & \multicolumn{2}{c}{\textbf{Quantized ANN}} & \multicolumn{2}{c}{\textbf{SDANN}} \\
            \cmidrule(lr){3-6}
   &            & MAC operations & Energy ($\mu J$) & AC operations & Energy ($\mu J$)   \\
            \midrule
            \multirow{2}{*}{\textbf{CIFAR10}}  & CIFAR10Net & \num{4.49e8}       & \num{103.27}    & \num{1.44e8}      & \num{4.32}\\
            ~                                  & ResNet-18   & \num{1.43e8}       & \num{32.89}    & \num{5.92e7}      & \num{1.78}\\
            \midrule
            \multirow{4}{*}{\textbf{ImageNet}} & VGG-16     & \num{9.49e9}       & \num{2182.7}    & \num{4.89e9}      & \num{146.7}  \\
            ~                                  & Tiny-VGG   & \num{4.15e8}       & \num{947.6}    & \num{3.27e8}      & \num{9.81}  \\
            ~                                  & ResNet-18  & \num{1.70e9}       & \num{391.0}    & \num{1.19e9}      & \num{35.7}  \\
            ~                                  & ResNet-34  & \num{3.55e9}       & \num{8165.0}    & \num{2.36e9}      & \num{70.8}  \\
            \bottomrule
    \end{tabular}

\end{table*}

\section{SDANN Models on Neuromorphic Hardware}\label{sec:impl}
\begin{figure*}[ht]
\centering 
    \includegraphics[width=\textwidth]{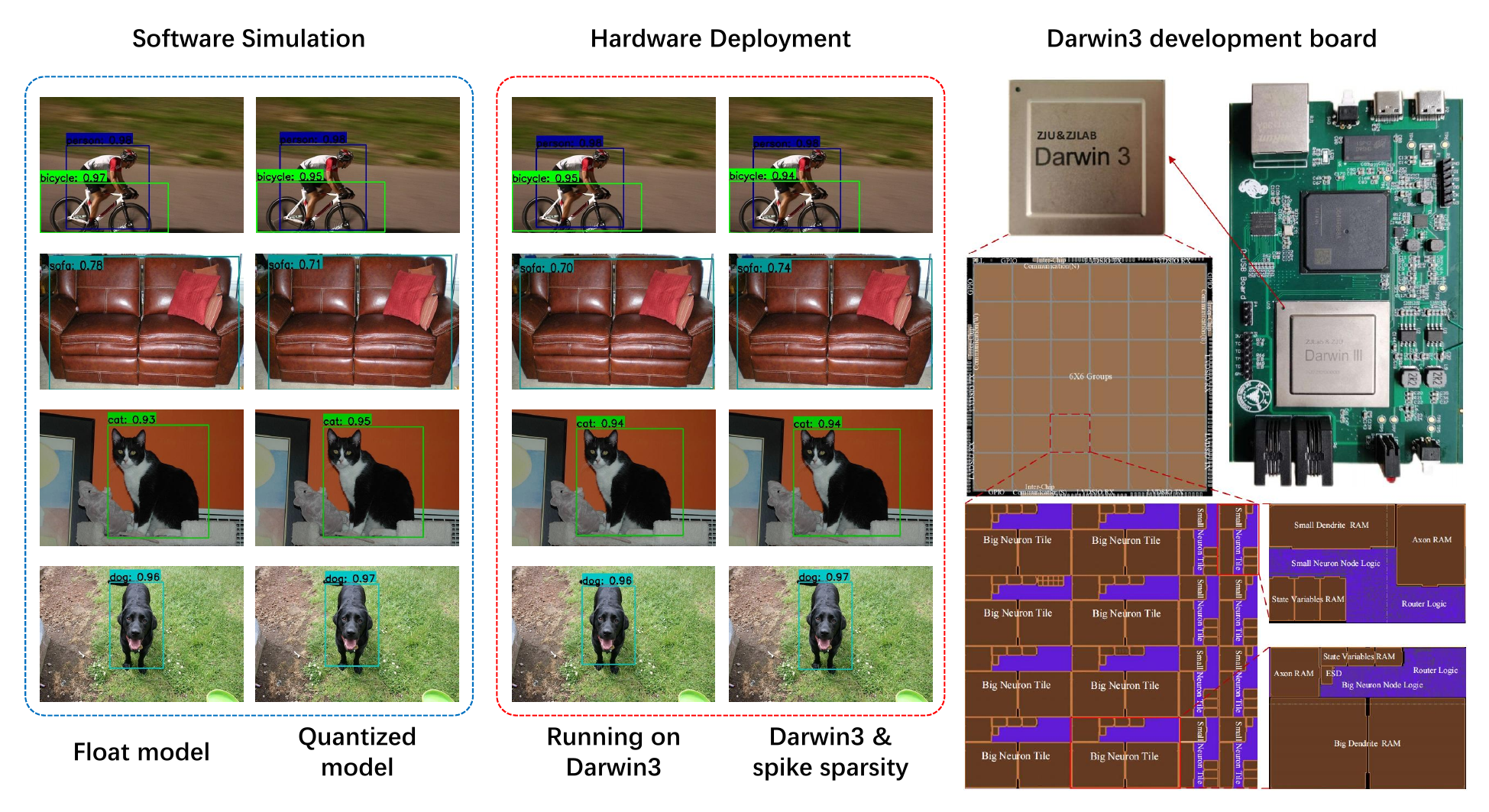} 
    \caption{Detection results comparison of Tiny-Yolo on the PASCAL VOC 2007 test set. From left to right: full-precision ANN, 8-bit quantized ANN, SDANN without spike sparsification, and SDANN with spike sparsification.} 
    \label{fig:visual}
\end{figure*}

In this section, we implement and evaluate SDANN models on real neuromorphic hardware.
We use Darwin3, a large-scale neuromorphic chip equipped with a neuron instruction set architecture, to validate SDANN and its internal components.
The first step is to compile the model structure, transform the weight and STEM into dendrites and neurons.
These dendrites and neurons are then deployed on Darwin3 hardware.
Input images are preprocessed and converted into spike trains, which are subsequently injected into the hardware.
The resulting output spike sequences indicate the model's prediction.

We deploy the Tiny-VGG and Tiny-YOLO networks on Darwin3 for image classification on ImageNet and object detection on VOC2007, respectively. To evaluate energy consumption, we randomly select 100 samples from each dataset and compute the average energy usage per sample.

\cref{tab:darwin3_energy} compares the inference energy cost per sample on Darwin3 and an NVIDIA GeForce RTX 3080 GPU. The Darwin3 hardware achieves significantly lower energy consumption than the GPU baseline. Furthermore, networks incorporating spike sparsity exhibit even lower energy efficiency on Darwin3, with the measured reduction falling within the same order of magnitude as the theoretically predicted SOP reduction from the spike sparsity method in \cref{tab:sparse coding}. The discrepancies are due to the circuit-level constraints of the Darwin3 chip used in our experiments.

\cref{tab:SNN on chip} presents a comparative analysis of spiking models performance reported on neuromorphic hardware in literature.
A key advantage of SDANN is that we can precisely and flawlessly map an ANN, as opposed to convert to SNN, to build up a spiking model for the neuromorphic hardware.
Freeing the implementation from ANN conversion can facilitate the spiking model deployment for more extensive networks and the execution of more intricate tasks on neuromorphic hardware.

The visual results of YOLOv1 are also presented in \cref{fig:visual}.
As demonstrated in the figure, the efficacy of our proposed method on neuromorphic hardware is commensurate with that of the full-precision floating-point model.
\begin{table}[h]
\centering
\caption{Energy Consumption Comparison Between GPU and Neuromorphic Hardware (with/without spike sparsity) per image sample.}
\label{tab:darwin3_energy}
\begin{tabular}{lccc}
\toprule
\multirow{2}{*}{\textbf{Network}} & \multirow{2}{*}{\textbf{GPU (mJ)}} & \multicolumn{2}{c}{\textbf{Neuromorphic Hardware (mJ)}} \\
\cmidrule(lr){3-4}
& & \textbf{without sparsity} & \textbf{with sparsity} \\
\midrule
Tiny-VGG        & 375.69       & 22.31            & 20.95(-6.11\%)            \\
Tiny-Yolo       & 234.44       & 23.80            & 21.71(-8.76\%)            \\
\bottomrule
\end{tabular}
\end{table}
\section{Conclusion and Prospect}\label{sec:discussion}
In this paper, we propose a novel framework, SDANN, which enables the direct deployment of a quantized ANN on neuromorphic hardware. To ensure hardware compatibility, we introduce scaled integration and bias calibration techniques that align intermediate computations with hardware constraints.
Experimental results demonstrate that our method achieves higher accuracy and fewer spikes compared to direct SNN training approaches based on surrogate gradients. Additionally, we implement two spike sparsification methods, RoT and DRLOs, to further reduce power consumption.
Ablation studies show that these methods effectively reduce the number of SOPs in SDANN with only a minor impact on accuracy.
This flexible and hardware-friendly sparsification strategy highlights the potential of neuromorphic hardware for low-power AI applications.
Beyond the software simulation, we successfully implement several SDANN models on real neuromorphic hardware.
This validates both the feasibility and practical utility of our proposed framework.

Based on this work, several promising directions remain for future improvement.
The current SDANN relies on uniform quantization of ANNs.
One potential avenue is to reduce the quantization bit width while preserving accuracy. This may be achieved by designing hardware-compatible models that approximate the behavior of non-uniform quantization, thereby enhancing overall performance.
Additionally, adopting more advanced activation functions for quantized ANNs, such as ReLU6, could help construct stronger base models, which in turn benefit the derived spiking implementations.

In summary, our work demonstrates that quantized ANNs—not only SNNs—can be directly mapped to neuromorphic hardware. In other words, the SDANN provides a {\it lower bound} for neuromorphic hardware as well as a new way to make full use of it. We believe this work will support and inspire further research toward energy-efficient AI systems.


\bibliographystyle{IEEEtran}
\bibliography{sn-bibliography}  

\begin{thebibliography}{10}
\providecommand{\url}[1]{#1}
\csname url@samestyle\endcsname
\providecommand{\newblock}{\relax}
\providecommand{\bibinfo}[2]{#2}
\providecommand{\BIBentrySTDinterwordspacing}{\spaceskip=0pt\relax}
\providecommand{\BIBentryALTinterwordstretchfactor}{4}
\providecommand{\BIBentryALTinterwordspacing}{\spaceskip=\fontdimen2\font plus
\BIBentryALTinterwordstretchfactor\fontdimen3\font minus \fontdimen4\font\relax}
\providecommand{\BIBforeignlanguage}[2]{{%
\expandafter\ifx\csname l@#1\endcsname\relax
\typeout{** WARNING: IEEEtran.bst: No hyphenation pattern has been}%
\typeout{** loaded for the language `#1'. Using the pattern for}%
\typeout{** the default language instead.}%
\else
\language=\csname l@#1\endcsname
\fi
#2}}
\providecommand{\BIBdecl}{\relax}
\BIBdecl

\bibitem{dl}
Y.~LeCun, Y.~Bengio, and G.~Hinton, ``Deep learning,'' \emph{Nature}, vol. 521, no. 7553, pp. 436--444, 2015.

\bibitem{de2023growing}
A.~de~Vries, ``The growing energy footprint of artificial intelligence,'' \emph{Joule}, vol.~7, no.~10, pp. 2191--2194, 2023.

\bibitem{roy2019towards}
K.~Roy, A.~Jaiswal, and P.~Panda, ``Towards spike-based machine intelligence with neuromorphic computing,'' \emph{Nature}, vol. 575, no. 7784, pp. 607--617, 2019.

\bibitem{truenorth}
P.~A. Merolla, J.~V. Arthur, R.~Alvarez-Icaza, A.~S. Cassidy, J.~Sawada, F.~Akopyan, B.~L. Jackson, N.~Imam, C.~Guo, Y.~Nakamura \emph{et~al.}, ``A million spiking-neuron integrated circuit with a scalable communication network and interface,'' \emph{Science}, vol. 345, no. 6197, pp. 668--673, 2014.

\bibitem{loihi}
M.~Davies, N.~Srinivasa, T.-H. Lin, G.~Chinya, Y.~Cao, S.~H. Choday, G.~Dimou, P.~Joshi, N.~Imam, S.~Jain \emph{et~al.}, ``Loihi: A neuromorphic manycore processor with on-chip learning,'' \emph{IEEE Micro}, vol.~38, no.~1, pp. 82--99, 2018.

\bibitem{darwin3}
D.~Ma, X.~Jin, S.~Sun, Y.~Li, X.~Wu, Y.~Hu, F.~Yang, H.~Tang, X.~Zhu, P.~Lin \emph{et~al.}, ``Darwin3: a large-scale neuromorphic chip with a novel isa and on-chip learning,'' \emph{National Science Review}, vol.~11, no.~5, p. nwae102, 2024.

\bibitem{sparsecoding}
\BIBentryALTinterwordspacing
P.~T.~P. Tang, T.-H. Lin, and M.~Davies, ``Sparse coding by spiking neural networks: Convergence theory and computational results,'' 2017. [Online]. Available: \url{https://arxiv.org/abs/1705.05475}
\BIBentrySTDinterwordspacing

\bibitem{lin2018programming}
C.-K. Lin, A.~Wild, G.~N. Chinya, Y.~Cao, M.~Davies, D.~M. Lavery, and H.~Wang, ``Programming spiking neural networks on intel's loihi,'' \emph{Computer}, vol.~51, pp. 52--61, 2018.

\bibitem{transformer}
A.~Vaswani, ``Attention is all you need,'' \emph{Advances in Neural Information Processing Systems}, 2017.

\bibitem{rueckauer2017conversion}
B.~Rueckauer, I.-A. Lungu, Y.~Hu, M.~Pfeiffer, and S.-C. Liu, ``Conversion of continuous-valued deep networks to efficient event-driven networks for image classification,'' \emph{Frontiers in neuroscience}, vol.~11, p. 682, 2017.

\bibitem{sengupta2019going}
A.~Sengupta, Y.~Ye, R.~Wang, C.~Liu, and K.~Roy, ``Going deeper in spiking neural networks: Vgg and residual architectures,'' \emph{Frontiers in neuroscience}, vol.~13, p.~95, 2019.

\bibitem{kim2020spiking}
S.~Kim, S.~Park, B.~Na, and S.~Yoon, ``Spiking-yolo: spiking neural network for energy-efficient object detection,'' in \emph{Proceedings of the AAAI conference on artificial intelligence}, vol.~34, 2020, pp. 11\,270--11\,277.

\bibitem{cao2015spiking}
Y.~Cao, Y.~Chen, and D.~Khosla, ``Spiking deep convolutional neural networks for energy-efficient object recognition,'' \emph{International Journal of Computer Vision}, vol. 113, pp. 54--66, 2015.

\bibitem{diehl2015fast}
P.~U. Diehl, D.~Neil, J.~Binas, M.~Cook, S.-C. Liu, and M.~Pfeiffer, ``Fast-classifying, high-accuracy spiking deep networks through weight and threshold balancing,'' in \emph{2015 International joint conference on neural networks (IJCNN)}, 2015, pp. 1--8.

\bibitem{han2020rmp}
B.~Han, G.~Srinivasan, and K.~Roy, ``Rmp-snn: Residual membrane potential neuron for enabling deeper high-accuracy and low-latency spiking neural network,'' in \emph{Proceedings of the IEEE/CVF conference on computer vision and pattern recognition}, 2020, pp. 13\,558--13\,567.

\bibitem{deng2021optimal}
S.~Deng and S.~Gu, ``Optimal conversion of conventional artificial neural networks to spiking neural networks,'' \emph{arXiv preprint arXiv:2103.00476}, 2021.

\bibitem{li2024error}
Y.~Li, S.~Deng, X.~Dong, and S.~Gu, ``Error-aware conversion from ann to snn via post-training parameter calibration,'' \emph{International Journal of Computer Vision}, pp. 1--24, 2024.

\bibitem{huawei_fp32}
W.~Li, H.~Chen, J.~Guo, Z.~Zhang, and Y.~Wang, ``Brain-inspired multilayer perceptron with spiking neurons,'' in \emph{2022 IEEE/CVF Conference on Computer Vision and Pattern Recognition (CVPR)}, 2022, pp. 773--783.

\bibitem{hu2023fast}
Y.~Hu, Q.~Zheng, X.~Jiang, and G.~Pan, ``Fast-snn: fast spiking neural network by converting quantized ann,'' \emph{IEEE Transactions on Pattern Analysis and Machine Intelligence}, 2023.

\bibitem{wu2018stbp}
Y.~Wu, L.~Deng, G.~Li, J.~Zhu, and L.~Shi, ``Spatio-temporal backpropagation for training high-performance spiking neural networks,'' \emph{Frontiers in neuroscience}, vol.~12, p. 331, 2018.

\bibitem{wu2019direct}
Y.~Wu, L.~Deng, G.~Li, J.~Zhu, Y.~Xie, and L.~Shi, ``Direct training for spiking neural networks: Faster, larger, better,'' in \emph{Proceedings of the AAAI conference on artificial intelligence}, vol.~33, 2019, pp. 1311--1318.

\bibitem{zheng2021going}
H.~Zheng, Y.~Wu, L.~Deng, Y.~Hu, and G.~Li, ``Going deeper with directly-trained larger spiking neural networks,'' in \emph{Proceedings of the AAAI conference on artificial intelligence}, vol.~35, 2021, pp. 11\,062--11\,070.

\bibitem{2022mlf}
L.~Feng, Q.~Liu, H.~Tang, D.~Ma, and G.~Pan, ``Multi-level firing with spiking ds-resnet: Enabling better and deeper directly-trained spiking neural networks,'' in \emph{Proceedings of the International Joint Conference on Artificial Intelligence}, 2022, pp. 2471--2477.

\bibitem{meng2023towards}
Q.~Meng, M.~Xiao, S.~Yan, Y.~Wang, Z.~Lin, and Z.-Q. Luo, ``Towards memory-and time-efficient backpropagation for training spiking neural networks,'' in \emph{Proceedings of the IEEE/CVF International Conference on Computer Vision}, 2023, pp. 6166--6176.

\bibitem{massa2020efficient}
R.~Massa, A.~Marchisio, M.~Martina, and M.~Shafique, ``An efficient spiking neural network for recognizing gestures with a dvs camera on the loihi neuromorphic processor,'' in \emph{2020 International Joint Conference on Neural Networks (IJCNN)}, 2020, pp. 1--9.

\bibitem{shrestha2021hardware}
A.~Shrestha, H.~Fang, D.~P. Rider, Z.~Mei, and Q.~Qiu, ``In-hardware learning of multilayer spiking neural networks on a neuromorphic processor,'' in \emph{2021 58th ACM/IEEE Design Automation Conference (DAC)}, 2021, pp. 367--372.

\bibitem{renner2024backpropagation}
A.~Renner, F.~Sheldon, A.~Zlotnik, L.~Tao, and A.~Sornborger, ``The backpropagation algorithm implemented on spiking neuromorphic hardware,'' \emph{Nature Communications}, vol.~15, no.~1, p. 9691, 2024.

\bibitem{frenkel202028}
C.~Frenkel, J.-D. Legat, and D.~Bol, ``A 28-nm convolutional neuromorphic processor enabling online learning with spike-based retinas,'' in \emph{2020 IEEE International Symposium on Circuits and Systems (ISCAS)}, 2020, pp. 1--5.

\bibitem{goltz2021fast}
J.~G{\"o}ltz, L.~Kriener, A.~Baumbach, S.~Billaudelle, O.~Breitwieser, B.~Cramer, D.~Dold, A.~F. Kungl, W.~Senn, J.~Schemmel \emph{et~al.}, ``Fast and energy-efficient neuromorphic deep learning with first-spike times,'' \emph{Nature machine intelligence}, vol.~3, no.~9, pp. 823--835, 2021.

\bibitem{spilger2023hxtorch}
P.~Spilger, E.~Arnold, L.~Blessing, C.~Mauch, C.~Pehle, E.~M{\"u}ller, and J.~Schemmel, ``hxtorch. snn: Machine-learning-inspired spiking neural network modeling on brainscales-2,'' in \emph{Proceedings of the 2023 Annual Neuro-Inspired Computational Elements Conference}, 2023, pp. 57--62.

\bibitem{huang2023efficient}
J.~Huang, F.~Kelber, B.~Vogginger, B.~Wu, F.~Kreutz, P.~Gerhards, D.~Scholz, K.~Knobloch, and C.~G. Mayr, ``Efficient algorithms for accelerating spiking neural networks on mac array of spinnaker 2,'' in \emph{2023 IEEE 5th International Conference on Artificial Intelligence Circuits and Systems (AICAS)}, 2023, pp. 1--5.

\bibitem{jacob2018quantization}
B.~Jacob, S.~Kligys, B.~Chen, M.~Zhu, M.~Tang, A.~Howard, H.~Adam, and D.~Kalenichenko, ``Quantization and training of neural networks for efficient integer-arithmetic-only inference,'' in \emph{Proceedings of the IEEE conference on computer vision and pattern recognition}, 2018, pp. 2704--2713.

\bibitem{horowitz20141}
M.~Horowitz, ``1.1 computing's energy problem (and what we can do about it),'' in \emph{2014 IEEE international solid-state circuits conference digest of technical papers (ISSCC)}, 2014, pp. 10--14.

\bibitem{liu2022spikeconverter}
F.~Liu, W.~Zhao, Y.~Chen, Z.~Wang, and L.~Jiang, ``Spikeconverter: An efficient conversion framework zipping the gap between artificial neural networks and spiking neural networks,'' in \emph{Proceedings of the AAAI Conference on Artificial Intelligence}, vol.~36, 2022, pp. 1692--1701.

\bibitem{pytorch}
\BIBentryALTinterwordspacing
A.~Paszke, S.~Gross, F.~Massa, A.~Lerer, J.~Bradbury, G.~Chanan, T.~Killeen, Z.~Lin, N.~Gimelshein, L.~Antiga, A.~Desmaison, A.~Kopf, E.~Yang, Z.~DeVito, M.~Raison, A.~Tejani, S.~Chilamkurthy, B.~Steiner, L.~Fang, J.~Bai, and S.~Chintala, ``Pytorch: An imperative style, high-performance deep learning library,'' in \emph{Advances in Neural Information Processing Systems}, H.~Wallach, H.~Larochelle, A.~Beygelzimer, F.~d\textquotesingle Alch\'{e}-Buc, E.~Fox, and R.~Garnett, Eds., vol.~32, 2019, pp. 8026--8037. [Online]. Available: \url{https://proceedings.neurips.cc/paper_files/paper/2019/file/bdbca288fee7f92f2bfa9f7012727740-Paper.pdf}
\BIBentrySTDinterwordspacing

\bibitem{cifar}
A.~Krizhevsky, ``Learning multiple layers of features from tiny images,'' in \emph{Proceedings of the International Conference on Machine Learning}, 2009.

\bibitem{imagenet}
J.~Deng, W.~Dong, R.~Socher, L.-J. Li, K.~Li, and L.~Fei-Fei, ``Imagenet: A large-scale hierarchical image database,'' \emph{2009 IEEE Conference on Computer Vision and Pattern Recognition (CVPR)}, pp. 248--255, 2009.

\bibitem{voc}
M.~Everingham, L.~van Gool, C.~Williams, J.~Winn, A.~Zisserman, Y.~Aytar, A.~Eslami, and A.~Sorokin, ``The pascal visual object classes homepage,'' \url{http://host.robots.ox.ac.uk/pascal/VOC/}, 2012.

\bibitem{yolo}
J.~Redmon, S.~Divvala, R.~Girshick, and A.~Farhadi, ``You only look once: Unified, real-time object detection,'' in \emph{Proceedings of the IEEE Conference on Computer Vision and Pattern Recognition (CVPR)}, 2016.

\bibitem{xu_qi_cvpr_kdsnn}
Q.~Xu, Y.~Li, J.~Shen, J.~K. Liu, H.~Tang, and G.~Pan, ``Constructing deep spiking neural networks from artificial neural networks with knowledge distillation,'' in \emph{2023 IEEE/CVF Conference on Computer Vision and Pattern Recognition (CVPR)}, 2023, pp. 7886--7895.

\bibitem{fang2021plif}
W.~Fang, Z.~Yu, Y.~Chen, T.~Masquelier, T.~Huang, and Y.~Tian, ``Incorporating learnable membrane time constant to enhance learning of spiking neural networks,'' in \emph{Proceedings of the IEEE/CVF international conference on computer vision}, 2021, pp. 2661--2671.

\end{thebibliography}

\end{document}